\documentclass{article}

\usepackage{arxiv}
\usepackage[round]{natbib}

\usepackage[utf8]{inputenc} 
\usepackage[T1]{fontenc}    
\usepackage{hyperref}       
\usepackage{url}            
\usepackage{booktabs}       
\usepackage{amsfonts}       
\usepackage{nicefrac}       
\usepackage{microtype}      
\usepackage{xcolor}         

\usepackage{microtype}
\usepackage{graphicx}
\usepackage{booktabs} 

\usepackage{microtype}
\usepackage{graphicx}
\usepackage{booktabs} 

\usepackage{natbib}
\usepackage{amssymb, amsmath,amsthm}
\usepackage{amsfonts}
\usepackage{algorithm}
\usepackage{algorithmic}
\usepackage{paralist}
\usepackage{multirow}
\usepackage{times}
\usepackage{wrapfig}

\usepackage{url,enumerate}
\usepackage{color,xcolor}
\usepackage{makeidx}  
\usepackage{amsmath,amssymb}
\usepackage{mathtools}
\usepackage[small, compact]{titlesec}
\usepackage{xspace}
\usepackage{epstopdf}
\usepackage{cite}
\usepackage{mathrsfs}
\usepackage{times}
\usepackage{enumerate}
\usepackage{color}
\usepackage{graphicx,epsfig}
\usepackage{amsmath,amssymb,xspace}
\usepackage{url}
\usepackage{hyperref}
\usepackage{bm}
\usepackage{bbm}
\usepackage{upgreek}
\usepackage{cleveref}
\usepackage{multirow}
\usepackage{ulem}
\usepackage{cancel}
\usepackage{subcaption}
\usepackage{dsfont}
\usepackage{hyperref}

\newtheorem{theorem}{Theorem}[section]

\newtheorem{lem}[theorem]{Lemma}
\newcommand{\ours}{NEST-H$\Gamma$P\xspace}
\newcommand{\gap}{$\Gamma$P\xspace}
\newcommand{\gaps}{$\Gamma$Ps\xspace}
\newcommand{\eg}{{\textit{e.g.},}\xspace}
\newcommand{\ie}{{\textit{i.e.},}\xspace}
\newcommand{\etc}{{\textit{etc}.}\xspace}

\renewcommand{\a}{{\bf a}}
\renewcommand{\b}{{\bf b}}

\renewcommand{\d}{{\rm d}}  

\newcommand{\f}{{\bf f}}
\newcommand{\g}{{\bf g}}

\newcommand{\bi}{{\bf i}}

\newcommand{\s}{{\bf s}}

\renewcommand{\u}{{\bf u}}

\newcommand{\x}{{\bf x}}
\newcommand{\y}{{\bf y}}
\newcommand{\z}{{\bf z}}

\newcommand{\Dcal}{\mathcal{D}}

\newcommand{\I}{{\bf I}}
\newcommand{\J}{{\bf J}}
\newcommand{\K}{{\bf K}}
\renewcommand{\L}{{\bf L}}
\newcommand{\Lcal}{{\mathcal{L}}}

\newcommand{\Bcal}{{\mathcal{B}}}

\newcommand{\N}{\mathcal{N}}  

\newcommand{\R}{{\bf R}}

\newcommand{\Scal}{{\mathcal{S}}}

\newcommand{\U}{{\bf U}}
\newcommand{\Ucal}{{\mathcal{U}}}

\newcommand{\Wcal}{{\mathcal{W}}}

\newcommand{\Ycal}{{\mathcal{Y}}}
\newcommand{\Z}{{\bf Z}}
\newcommand{\Zcal}{{\mathcal{Z}}}

\newcommand{\bphi}{\boldsymbol{\phi}}

\newcommand{\bbeta}{\boldsymbol{\beta}}

\newcommand{\bnu}{\boldsymbol{\nu}}

\newcommand{\btheta}{\boldsymbol{\theta}}

\newcommand{\bomega}{\boldsymbol{\omega}}

\newcommand{\talpha}{\widetilde{\alpha}}
\newcommand{\tbeta}{{\widetilde{\bbeta}}}

\newcommand{\tomega}{{\widetilde{\bomega}}}
\newcommand{\ttomega}{{\widetilde{\omega}}}

\newcommand{\hA}{{\widehat{A}}}

\newcommand{\bmu}{\boldsymbol{\mu}}
\newcommand{\1}{{\bf 1}}
\newcommand{\0}{{\bf 0}}

\newcommand{\ben}{\begin{enumerate}}
\newcommand{\een}{\end{enumerate}}

\newcommand{\EE}{\mathbb{E}}
\newcommand{\cmt}[1]{}

\title{Nonparametric Sparse Tensor Factorization with Hierarchical Gamma Processes}

\author{ Conor Tillinghast\\
	Department of Mathematics\\
	University of Utah\\
	Salt Lake City, UT 84112 \\
	\texttt{ctilling@math.utah.edu} \\
	\And
	Zheng Wang\\
	School of Computing\\
	University of Utah\\
	Salt Lake City, UT 84112 \\
	\texttt{wzhut@cs.utah.edu} \\
	\And
	Shandian Zhe\\
	School of Computing\\
	University of Utah\\
	Salt Lake City, UT 84112 \\
	\texttt{ zhe@cs.utah.edu} \\
}
\date{}
\begin{document}
\maketitle
\begin{abstract} 
We propose a nonparametric factorization approach for sparsely observed tensors.  The sparsity does not mean zero-valued entries are massive or dominated. Rather, it implies the observed entries are very few, and even fewer with the growth of the tensor; this is ubiquitous in practice. Compared with the existent works, our model not only leverages the structural information underlying the observed entry indices, but also provides extra interpretability and flexibility --- it can simultaneously estimate a set of location factors about the intrinsic properties of the tensor nodes, and another set of sociability factors reflecting their extrovert activity in interacting with others; users are free to choose a trade-off between the two types of factors. Specifically, we use hierarchical Gamma processes and Poisson random measures to construct a tensor-valued process, which can freely sample the two types of factors to generate tensors and always guarantees an asymptotic sparsity.  We then normalize the tensor process to obtain hierarchical Dirichlet processes to sample each observed entry index, and use a Gaussian process to sample  the entry value as a nonlinear function of the factors, so as to capture both the sparse structure properties and complex node relationships. For efficient inference, we use Dirichlet process properties over finite sample partitions, density transformations, and random features to develop an stochastic variational estimation algorithm. We demonstrate the advantage of our method in several benchmark datasets. 
\end{abstract}	
\section{Introduction}
\vspace{-0.1in}
Tensor factorization is a popular tool to analyze multiway data, such as purchase records among \textit{(users, products, sellers)}  in online shopping. While numerous algorithms have been proposed, \eg \citep{Tucker66,Harshman70parafac,Chu09ptucker,kang2012gigatensor}, most of these methods are essentially modeling dense tensors.  For instance, many of them~\citep{Kolda09TensorReview,kang2012gigatensor,choi2014dfacto,xu2012infinite} use tensor and matrix algebras~\citep{kolda2006multilinear} to update the latent factors and hence demand all the tensor entries be observed (although the majority are often assumed to be zero-valued). While other methods~\citep{RaiDunson2014,zhe2016distributed,du2018probabilistic}, especially with a Bayesian formulation, can perform element-wise factorization to handle a small subset of observed entries,  from modeling perspective, they are equivalent to sampling a full tensor first and then marginalizing out the unobserved entry values. One might consider adding a Bernoulli likelihood (or a thresholding function) to model the presence of each entry. However, this still leads to dense tensors. In theory, these models are random function prior models~\citep{LloydOGR12randomgraph}, and the generated mask tensors are exchangeable arrays. According to the Aldous-Hoover theorem~\citep{aldous1981representations,hoover1979relations}, these tensors are ``either trivially empty or dense''. That is,  the present entries (\ie with mask one) increases linearly with the size of the tensor.

However, in practice, tensors are often sparsely observed. The ratio of observed entries is very small, \eg 0.01\%.\cmt{, slightly above zero  (\eg 0.01\%).} This ratio can even decrease with more nodes added. For example, in online shopping, with the growth of users, products and sellers, the number of actual purchases --- \ie the present entries --- while growing, can take an even smaller percentage of all possible purchases (entries), \ie all (user, product, seller) combinations, because the latter grows much faster. Hence, the (asymptotic) sparse nature in these real-world applications implies that the existent dense models are misspecified.  Note that the sparsity here does not refer to that the zero-valued entries are massive or dominated~\citep{choi2014dfacto,hu2015zero}

To address the model misspecification, \citet{tillinghast2021nonparametric} recently used Dirichlet processes (DPs) (normalized Gamma processes) and GEM distributions to build two asymptotically sparse tensor models, which have shown significant advantages over the popular dense models in entry value and index (link) prediction. However, their model structures severely restrict the type of factors to be learned --- either all the factors are sociability factors  to account for the generation of the interactions, or only one factor is the sociability factor and the remaining are location factors about the intrinsic properties of the tensor nodes. The former misses the chance to capture the intrinsic node properties while the latter can be insufficient to estimate the sociabilities.  Accordingly, both models can limit the flexibly of the factor estimation and lower the interpretability.

To address these limitations, we propose  a new nonparametric factorization model, which not only guarantees an asymptotic sparsity, but also provides more flexibility and interpretability. Our model can estimate an arbitrary number of sociability and location factors so as to fully capture the two types of tensor node properties.   Specifically, we use hierarchical Gamma processes (H\gap) to construct a summation of product measures, witch which we construct a tensor-variate process. We show that the proportion of the sampled entries is asymptotically tending to zero with the growth of tensor size. More important, the \gaps in the top level enable the sharing of locations across the  low-level \gap{s}. As a result,  for each node, our model can sample not only an arbitrary number of location factors representing the node's intrinsic properties, but also an arbitrary number of sociability factors accounting for its extrovert interaction activity in different communities. Hence, the results can provide more interpretability and users are flexible to select a trade-off. Next, for convenient modeling and inference, we normalize our tensor process to obtain hierarchical Dirichlet processes so as to sample the observed entry indices; we then use a Gaussian process to sample the entry value as a nonlinear function of the latent factors of the associated nodes so as to embed both the sparse structural knowledge and complex relationships between the tensor nodes. For efficient inference, we use the Dirichlet process property over  finite, non-overlapping sample space partitions to aggregate the sociabilities of non-active nodes, probability density transformations, and  random Fourier features to develop a scalable stochastic variational estimation algorithm. 

For evaluation, we first ran simulations to confirm that our method can indeed generate increasingly sparser tensors. In three real-world applications, we examined the performance in predicting missing entry indices and entry values. In both tasks, our approach nearly always obtains the best performance, as compared with the state-of-the-part dense models and \citep{tillinghast2021nonparametric}. The factors estimated by our method reveal clear, interesting clustering patterns while those estimated from the popular dense models cannot. 

\vspace{-0.1in}
\section{Background} \label{sect:bk}
\vspace{-0.1in}
\noindent\textbf{Notation and Settings.} We denote a full $K$-mode tensor by $\Ycal \in \mathbb{R}^{D_1 \times \ldots \times D_K}$, where mode $k$ includes $D_k$ nodes. Each entry is indexed by $\bi = (i_1, \ldots, i_K)$ and the entry value by $y_{\bi}$.  Practical tensors are often very sparse. That means most entries of $\Ycal$ are nonexistent. Given a small set of present entries, $\Dcal = \{(\bi_1, y_{\bi_1}), \ldots, (\bi_N, y_{\bi_N}) \}$ where $N\ll \prod_{k=1}^D D_k$, we aim to estimate a set of factors $\u_k^j$ to represent each node $j$ in  mode $k$. We denote by $\Ucal = \{\U^1, \ldots, \U^K\}$ all the factors, where each $\U^k = [\u^k_1, \ldots, \u^k_{D_k}]^\top$.  

\noindent \textbf{Dense Tensor Models.} The classical Tucker~\citep{Tucker66} and CANDECOMP/PARAFAC (CP) decomposition~\citep{Harshman70parafac}  require  $\Ycal$ to be fully observed. The Tucker decomposition assumes $\Ycal =  \Wcal \times_1 \U^{1} \times_2 \ldots \times_K \U^{K}$
where  $\Wcal \in \mathbb{R}^{r_1 \times \ldots \times r_K}$ is a parametric core tenor  and $\times_k$ is the  tensor-matrix product at mode $k$~\citep{kolda2006multilinear}. CP decomposition is  a special case of Tucker decomposition where $\Wcal$ is restricted to be diagonal.  While many methods can perform element-wise factorization over a  subset of observations, \eg~\citep{zhe2016distributed,raiscalable,schein2015bayesian,Schein:2016:BPT:3045390.3045686,zhe2018stochastic}, from the Bayesian perspective, these methods can be viewed as first sampling all the entry values of $\Ycal$ given the factors $\Ucal$, and then marginalize out the unobserved ones. In other words, they never consider the sparsity of the observed entry indices (links). Therefore, these methods are in essence still modeling a dense tensor. In theory, this has been justified by the random function prior models~\citep{LloydOGR12randomgraph}, where a Bernoulli distribution is used to sample the presence (or existence) of each entry given the factors, $p(\Zcal|\Ucal)= \prod_{\bi} p(z_{\bi}|\Ucal) = \prod_\bi \text{Bern}(z_{\bi}|f(\x_{\bi}))$, where $\Zcal$ is a mask tensor, $z_\bi = 1$ means the entry $\bi$ is generated, and $z_\bi =0$ means $\bi$ is nonexistent, $f$ is a function of latent factors, \eg element-wise CP/Tucker form or a Gaussian process (GP). The mask tensor $\Zcal$ is known to be exchangeable~\citep{LloydOGR12randomgraph}.  According to the Aldous-Hoover theorem~\citep{aldous1981representations,hoover1979relations}, the asymptotic\footnote{by asymptotic, it means when we sample a sequence of $\Zcal$'s, with increasing sizes (till infinity).} proportion of the present entries (\ie the entries with mask 1) is  $\eta = \int \text{Bern}(z_{\bi}|f(\x_{\bi})) \d p(\Ucal)$. This implies the sampled tensors are either empty ($\eta = 0$) or dense ($\eta>0$) almost surely, \ie the number of sampled entries grows linearly with the tensor size ($\Theta(\eta \prod_{k=1}^K D_K)$).

\noindent \textbf{Sparse Tensor Model}. Recently, \citet{tillinghast2021nonparametric} proposed nonparametric models that guarantee to generate asymptotically sparse tensors, \ie  the proportion of sampled entries tend to zero with the growth of the tensor size ($o(\prod_{k=1}^K D_K)$). The idea is to use Gamma processes (\gaps) to construct a tensor-variate random process. This is equivalent to sampling a Dirichlet process~\citep{ferguson1973bayesian} (normalized \gap) for each mode $k$, and combine the weights of each node  to sample the present entries, 
\begin{align}
	\text{DP}^k = \sum\nolimits_{j} w_j^k\delta_{\btheta^k_j}, \;\;\; p(\bi_n) = \prod\nolimits_k w_{i_{nk}}^k, \label{eq:nest-1}
\end{align}
where $1\le k \le K$, $1 \le n \le N$, $w_j^k$ and $\btheta^k_j$ are the DP weights and locations for node $j$ in mode $k$, which naturally represent the sociability and location factors. The sociability factor $w_j^k$ represents the activity of the node in interacting with others to generate the entry index, \ie link. The location factors naturally represents the intrinsic properties of the node. Then $w_j^k$ and $\btheta^k_j$ are jointly used to the sample the value of the observed entries $y_{\bi_n}$. To incorporate multiple sociability factors, \citet{tillinghast2021nonparametric} proposed a second model that draws multiple weights for each node via the  GEM distribution~\citep{griffiths1980lines,engen1975note,mccloskey1965model}. Note that GEM essentially samples DP weights (no locations). These weights are integrated to sample both the observed entry indices and entry values. 
 
\vspace{-0.05in}
\section{Model}
\vspace{-0.1in}
Despite being successful, the popular dense tensor models can be misspecified for the sparse data that are ubiquitous in practice. Although the work of \citep{tillinghast2021nonparametric} have overcome this issue, their model structures severely restrict the type of the factors to be estimated. From \eqref{eq:nest-1}, we can only estimate one sociability factor for each node to account for their complex interactions (\ie links), which can be quite insufficient. However, to incorporate multiple sociability factors, we have to use GEM distributions and drop the location factors (the second model).  Accordingly, we cannot capture any intrinsic properties of the nodes. Note that we cannot simply draw multiple DPs, because the indices of the nodes in different DPs are not necessarily the same and we are unable to align their location factors. 

To address these limitations, we propose a new sparse tensor factorization model, which not only guarantees the asymptotic sparsity in tensor generation, but also is free to estimate an arbitrary number of sociability and location factors so as to fully capture the extrovert and intrinsic properties of the tensor nodes. Accordingly, our model can provide more interpretability and additional flexibly to select the trade-off between the two types of the factors (\eg cross-validation). 
\vspace{-0.1in}
\subsection{H$\mathbf{\Gamma}$P Based Sparse Tensor-Variate Process}
\vspace{-0.1in}
Specifically, we use hierarchical Gamma processes (H\gaps) to construct a sparse tensor-variate process. At the top level, we sample a Gamma process (\gap)~\citep{hougaard1986survival} in each mode $k$, 
 \begin{align}
 	L^{\alpha}_k & \sim \Gamma\text{P}(\lambda_\alpha)  \;\; (1 \le k \le K), \label{eq:gap-1}
 \end{align}
where $\lambda_\alpha$ is a Lebesgue base measure restricted to $[0, \alpha]^{R_1}  = \underbrace{[0, \alpha] \times \ldots \times [0, \alpha]}_{R_1 \text{times}} (\alpha > 0)$. Accordingly, $L^\alpha_k$ is a discrete measure, 
\begin{align}
L^{\alpha}_k &= \sum\nolimits_{j=1}^\infty w^\alpha_{kj} \cdot \delta_{\btheta^k_j}
\end{align}
where $\delta_{[\cdot]}$ is the Dirac Delta measure, $\{w^\alpha_{kj}>0\}$ and $\{\btheta^k_j\in[0, \alpha]^{R_1}\}$ are the weights and locations, which correspond to an infinite number of tensor nodes in mode $k$. We refer to each $\btheta^k_j$  as $R_1$ location factors to represent the intrinsic properties of node $j$ in mode $k$. In the second level, we use $L_k^\alpha$ as the base measure to sample $R_2$ Gamma processes in each mode $k$, 
\begin{align}
	W^\alpha_{k,r} \mid L^\alpha_k \sim \Gamma\text{P}(L^\alpha_k), \;\;  W^\alpha_{k,r} = \sum\nolimits_{j=1}^\infty v^\alpha_{krj} \cdot \delta_{\btheta^k_j}, \label{eq:gap-2}
\end{align}
where $1 \le r \le R_2$. 
Note that due to the common base measure $L^\alpha_k$, the locations are shared across all $\{W^\alpha_{k,r}\}_{r=1}^{R_2}$. 
Hence,  for each node $j$ in mode $k$, we not only generate  $R_1$ location factors $\btheta_j^k$, but also $R_2$ weights, $\{v^\alpha_{krj}\}_{r=1}^{R_2}$. We refer to the normalized weights as  sociability  factors, since we will use them to sample the observed entry indices (links). 
More detailed discussions will be given in Sec. \ref{sect:finite-model}. 
Next, we use the H\gaps to construct a product measure sum as the rate (or mean) measure to sample a Poisson random measure (PRM)~\citep{kingman1992poisson}, which represents the sampled tensor entries, 
\begin{align}
	A &=  \sum_{r=1}^{R_2} W^\alpha_{1,r} \times \ldots \times W^\alpha_{K,r}, \notag \\
	T &\mid \{W^\alpha_{k,r}\}_{1\le k \le K, 1 \le r \le R_2} \sim  \text{PRM}(A). \label{eq:ppp}
\end{align}
Accordingly, $T$ has the following form, $T = \sum_{\bi \in \Scal} c_{\bi} \cdot \delta_{(\btheta^1_{i_1}, \ldots, \btheta^K_{i_K})}$, 
where each point represents an entry index, $\Scal$ is the set of all the sampled points in the PRM, \ie  entry indices, $c_{\bi}>0$ is the count of the point $\bi$, and $(\btheta^1_{i_1}, \ldots, \btheta^K_{i_K})$ is the location of that point.  

In a nut shell, the H\gaps sample an infinite number of nodes in each mode, then the PRM picks the nodes from each mode to generate the entries (\ie links). 
\begin{lem}
	For any fixed $\alpha>0$, the number of sampled entries $N^\alpha=|\Scal|$ via \eqref{eq:gap-1}\eqref{eq:gap-2}\eqref{eq:ppp} is finite almost surely. When $\alpha \rightarrow \infty$, $N^\alpha \rightarrow \infty$ a.s.
	\label{lem:1}
\end{lem}
We leave the proof in the Appendix.
To examine the sparsity, we are interested in the active nodes that show up in the sampled entry indices $\Scal$. For example, if an entry $(1, 3, 5)$ is sampled, then node 1, 3, 5 in mode 1, 2, 3 are active nodes. Denote by $D_k^\alpha$ the number of (distinct) active nodes in mode $k$. If we use these active nodes to construct a full tensor, which we refer to as the active tensor, the size will be $\prod_{k=1}^K D_k^\alpha$. The asymptotic sparsity means the proportion of the present entries in the active tensor is small, and the increase of present entries is much slower than the growth of the active tensor size. This is guaranteed by 
\begin{lem}
	\cmt{If the tensor entries are sampled by the model defined in \eqref{eq:gap-1}\eqref{eq:gap-2}\eqref{eq:ppp}, we have}  $N^\alpha = o(\prod _{k=1}^K D^\alpha_k)$ almost surely as $\alpha \rightarrow \infty$, \ie  
	$\underset{\alpha \rightarrow \infty}{\lim } \frac{N^\alpha}{\prod _{k=1}^K D^\alpha_k} = 0 \;\;\;a.s. $
	\label{lem:2}
\end{lem}
The proof framework is similar to that in \citep{tillinghast2021nonparametric}, and we leave the details in the Appendix. 
We refer to the model defined by \eqref{eq:gap-1}\eqref{eq:gap-2}\eqref{eq:ppp} as our Sparse Tensor-variate Process (STP). 



\vspace{-0.05in}
\subsection{Projection on Finite Data}\label{sect:finite-model}
\vspace{-0.05in}
Directly using the STP for modeling and inference is inconvenient, because the infinite discrete measures of \gaps and PRMs are computationally challenging. On the other hand, in practice, we always observe a finite number of entries, $\Dcal = \{(\bi_1, y_1), \ldots, (\bi_N, y_N)\}$. Hence, we can  use the standard PRM construction~\citep{kingman1992poisson}  to equivalently model the sampling procedure. That is, we normalize the rate measure $A = \sum_{r=1}^{R_2} W^\alpha_{1r} \times \ldots \times W^\alpha_{Kr}$ in \eqref{eq:ppp} to obtain a probability measure, and use it to sample  $N$ entry indices (points) independently. To normalize $A$, we need to first normalize each \gap $W^\alpha_{kr} (1 \le k \le K, 1 \le r \le R_2 )$, which gives a Dirichlet process  (DP)~\citep{ferguson1973bayesian} with the base measure as the normalized base measure of $W^\alpha_{kr}$. Since the base measure of $W^\alpha_{kr}$ is another \gap, $L^\alpha_k$ (see \eqref{eq:gap-1}), its normalization is a DP again. Hence, we obtain a set of hierarchical Dirichlet processes (HDPs)~\citep{teh2006hierarchical}, 
\begin{align}
	G_k &\sim \text{DP}\big(\talpha, \text{Uniform}([0, \alpha]^{R_1}) \big), \notag \\
	H_r^k & \sim \text{DP}(\gamma_{r}^k,  G_k) \label{eq:hdp}
\end{align}
where $1 \le k \le K$, $1 \le r \le R_2$, $\talpha = \alpha^{R_1}$, and $\gamma_{r}^k \sim \text{Gamma}(\cdot|1, \talpha)$. \cmt{See more detailed explanation in the appendix.} Then the normalized $A$ is $\hA = \frac{1}{R_2} \sum_{r=1}^{R_2} H^1_{r} \times \ldots \times H^K_{r}$. The sampling of the $N$ observed entry indices is as follows. We first sample each $G_k$ and $H_r^k$ from \eqref{eq:hdp}, which gives 
\begin{align}
	G_k = \sum\nolimits_{j=1}^\infty \beta^k_j \cdot \delta_{\btheta^k_j}, \;\;\; H_r^k = \sum\nolimits_{j=1}^\infty \omega^k_{rj}\cdot  \delta_{\btheta^k_j},  \label{eq:hdp-sample}
\end{align}
where each location $\btheta^k_j$ is independently sampled from $\text{Uniform}([0, \alpha]^{R_1})$. Then we obtain the sample of the normalized rate measure as 
\begin{align}
\hA = \sum\nolimits_{\bi=(1, \ldots, 1)}^{(\infty, \ldots, \infty)}  w_{\bi} \cdot \delta_{(\btheta^1_{i_1}, \ldots, \btheta^K_{i_K})}, \label{eq:entry-dist}
\end{align}
where $w_{\bi} = \frac{1}{R_2} \sum_{r=1}^{R_2} \prod_{k=1}^K \omega_{ri_{k}}^k$
which is essentially a probability measure over all possible tensor entries, \ie $\sum_{\bi} w_{\bi} = 1$. We then sample each observed entry $\bi_n \sim \hA$, and hence the probability is 
\begin{align}
	p(\Scal) = \prod\nolimits_{n=1}^N p(\bi_n) = \prod\nolimits_{n=1}^N w_{\bi_n}.  \label{eq:edge-prob}
\end{align}

Now, it can be seen clearly that in \eqref{eq:hdp}, for each node $j$ in mode $k$, we sample the location $\btheta_j^k$ and a set of weights $\bnu^k_j = [\omega^k_{1j}, \ldots, \omega^k_{R_2j}]$ from $R_2$ DPs. From \eqref{eq:entry-dist}, we can see that these weights represent the activity of the node interacting with other nodes. Each weight naturally corresponds to the activity in one community/group. Hence, we refer to $\bnu^k_j$ as the sociability factors of the node in $R_2$ overlapping communities. The location factors $\btheta_j^k$ naturally represent the node's intrinsic properties. Thanks to the HDP (H\gap) structure, the locations can be shared in an arbitrary number of low-level DPs (\gaps). That means, we are free to choose the number of sociability factors and location factors. Hence, it provides not only more interpretability  but also an extra flexibility to select their trade-off. 

Given the sampled entries $\Scal$, we then sample the entry values $\y = [y_{\bi_1}, \ldots, y_{\bi_N}]$. In this work, we mainly consider continuous values. It is straightforward to extend our model for other value types. We sample each value from a Gaussian noise model,  
\begin{align}
	p(y_{\bi_n}|\Scal) = \N(y_{\bi_n}|f(\x_{\bi_n}), \sigma^2),  \label{eq:value-prob}
\end{align}
where $\x_\bi = [\u^1_{i_{n1}}; \ldots ; \u^K_{i_{nK}}]$ are the factors of all the nodes associated with entry $\bi_n$,  each $\u^k_{i_{nk}} = [\btheta^k_{i_{nk}}; \bnu^k_{i_{nk}}]$, \ie including $R_1$ location and $R_2$ sociability factors, and $f(\cdot)$ is a latent factorization function. To flexibly estimate $f(\cdot)$ so as to capture the complex relationships between the tensor nodes (in terms of their factor representations), we assign a Gaussian process~\citep{Rasmussen06GP} prior over $f(\cdot)$. Accordingly, the function values $\f = [f(\x_{\bi_1}), \ldots, f(\x_{\bi_N})]$ follow a joint Gaussian distribution, $p(\f) = \N(\f|\0, \K)$,
where $\K$ is an $N \times N$ kernel (covariance) matrix, each element $[\K]_{st} = \kappa(\x_{\bi_s}, \x_{\bi_t})$, and $\kappa(\cdot, \cdot)$ is a kernel function. 

From \eqref{eq:edge-prob} and \eqref{eq:value-prob}, we can see via coupling the HDPs and GP, both the structural knowledge in sparse entry indices and complex relationships of the tensor nodes can be captured and encoded into the sociability and location factors.   

\vspace{-0.1in}
\section{Algorithm}\label{sec:alg}
\vspace{-0.1in}
The estimation of our model is still challenging. First, we have to deal with infinite discrete measures  (see \eqref{eq:hdp-sample}) from HDPs. Second, the exact GP prior includes an $N \times N$ covariance matrix. When $N$ is large, the computation is very costly or even infeasible. To address these issues, we use the DP property over sample partitions~\citep{ferguson1973bayesian}, probability density transformations, and random Fourier features~\citep{rahimi2007random,lazaro2010sparse} to develop  an efficient, scalable variational learning algorithm. 

Specifically, since we only need to estimate the latent factors for finite active nodes, \ie the nodes appearing in the observed entries (the infinite inactive nodes do not have any observed data), we can marginalize \eqref{eq:hdp-sample} into finite measures. Suppose we have $D_k$ active nodes in each mode $k$. Without loss of generality, we can index these nodes by $1, \ldots, D_k$, and the remaining inactive nodes by $D_{k+1}, \ldots, \infty$.  In the first level, for each $G_k$, we consider $\bbeta^k = [\beta^k_1, \ldots, \beta^k_{D_k}, \beta^k_{>D_k}]$ where $\beta^k_{>D_k} = \sum_{j=D_k+1}^\infty \beta^k_{j}$, \ie the DP weights for every active node and the aggregated weight of all the inactive nodes. In the second level,  we partition the sample space of each $H^k_{r}$ into $D_k+1$ subsets, $\{\{\btheta_1^k\}, \ldots, \{\btheta_{D_k}^k\}, \{\text{Others}\}\}$. Since $H^k_r \sim \text{DP}(\gamma_r^k, G_k)$,  its measures of the $K+1$ subsets will jointly follow a Dirichlet distribution. The values of these measures are $\bomega^k_r = \{\omega_{r1}^k, \ldots, \omega_{rD_k}^k, \omega_{r>D_k}^k\}$, where $\omega^k_{r>D_k} = \sum_{j=D_k+1}^\infty \omega^k_{rj}$, namely, the sociability of each active node and the aggregated sociability of all the inactive nodes.  The Dirichlet distribution is parameterized by  $[\gamma_r^k G_k(\{\btheta_1^k\}), \ldots, \gamma_r^k G_k(\{\btheta_{D_k}^k\}), \gamma_r^k G_k(\{\text{Others}\})] = \gamma_r^k \bbeta^k$.   Therefore, we have
\begin{align}
	&p(\bomega^k_r|\gamma^k_r, \bbeta^k) = \text{Dir}(\bomega^k_r|\gamma^k_r\bbeta^k) \notag \\
	&= \frac{\Gamma(\gamma_r^k)}{\prod_{j}\Gamma(\gamma_r^k \beta^k_j)} \prod\nolimits_{j} \left(\omega^k_{rj}\right)^{\gamma_r^k\beta^k_j - 1},  \label{eq:dir-dist}
\end{align}
where $\Gamma(\cdot)$ is a Gamma function. Given the sociabilities of the active nodes, we have already been able to sample the observed entry indices from \eqref{eq:edge-prob}. The remaining problem is how to obtain the prior of  $\bbeta^k$ for each $G_k$.  From the stick-breaking construction~\citep{ishwaran2001gibbs}, we have $\xi^k_j \sim \text{Beta}(1, \talpha), \beta^k_j = \xi^k_j\prod\nolimits_{t=1}^{j-1} (1-\xi^k_t)\; (1 \le j \le \infty)$, 
from which we can obtain a reverse mapping,  
\begin{align}
(\xi^k_1, \ldots, \xi^k_{D_k}) = (\frac{\beta^k_1}{\Lambda^k_1}, \ldots, \frac{\beta^k_{D_k}}{\Lambda^k_{D_k}}),
\end{align}
where $\Lambda^k_j = 1 - \sum_{t=1}^{j-1} \beta^k_t = \prod_{t=1}^{j-1}(1 - \xi^k_t)$. Therefore, we can use probability density  transformation to obtain the prior distribution of $\bbeta^k$. Since  each $\Lambda_j^k$ is only determined by $\{\beta_1^k, \ldots, \beta_{j-1}^k\}$,  the Jacobian $\J$ is a lower-triangular matrix and $|\J| = \prod_{j=1}^{D_k} \frac{1}{\Lambda^k_j}$. Accordingly, we can derive  
\begin{align}
	p(\bbeta^k) = \prod\nolimits_{j=1}^{D_k} \text{Beta}(\frac{\beta^k_j}{\Lambda^k_j}|1, \talpha) \frac{1}{\Lambda^k_j}. \label{eq:beta-dist}
\end{align}
Note that  since $\beta^k_{>D_k} = 1 - \sum_{j=1}^{D_k} \beta^k_j$, it does not need to be explicitly computed in the prior. To conveniently estimate each $\bbeta^k$ and $\bomega^k_{r}$ and to avoid incorporating additional constraints (nonnegative, summation is one), we parameterize $\bbeta^k = \text{softmax}(\tbeta^k)$ and $\bomega^k_r = \text{softmax}(\tomega^k_r)$ and estimate the free parameters $\{\tbeta^k, \tomega^k_r\}$ instead. 

Next, to scale to a large number of observed entries, we use random Fourier features to construct a sparse GP approximation~\citep{rahimi2007random,lazaro2010sparse}. Specifically, according to Bochner's theorem~\citep{rudin1962fourier}, any stationary kernel $\kappa(\b_1, \b_2) = \kappa(\b_1 -\b_2)$ can be viewed as an expectation, $\kappa(\b_1 -\b_2) = \kappa(\0)\EE_{p(\z)}[e^{i \z^\top \a_1} \big(e^{i  \z^\top \a_2 }\big)^\dagger]$, where $\dagger$ is the complex conjugate, $p(\z)=\N(\z|0, \tau\I)$ is computed from the Fourier transform of $\kappa(\cdot)$. Hence, we can draw $M$ frequencies $\Z = \{\z_1, \ldots, \z_M\}$ from $p(\z)$ to construct an Monte-Carlo approximation,  $\kappa(\b_1,\b_2) \approx \frac{\kappa(\0)}{M}\sum_{m=1}^M  e^{i \s_m^\top \b_1} \big(e^{i s_m^\top \b_2 }\big)^\dagger  = \frac{\kappa(\0)}{M} \bphi(\b_1)^\top \bphi(\b_2)$, where $$\bphi(\b) = [\cos(\z_1^\top \b), \sin(\z_1^\top \b), \ldots, \cos(\z_M^\top \b), \sin(\z_M^\top \b)].$$
Here, $\bphi(\cdot)$ is a $2M$ dimensional nonlinear feature mapping, which is called random Fourier features.  We can then use a Bayesian linear regression model over the random Fourier features to approximate the GP. Specifically, we use the RBF kernel, $\kappa(\b_1, \b_2) = \exp(-\frac{1}{2} \tau \|\b_1 - \b_2\|^2)$. The corresponding frequency distribution $p(\z) = \N(\z|\0, \tau\I)$.  We sample $M$ frequencies, $\Z = \{\z_1, \ldots, \z_M\}$, from  $p(\Z) = \prod_{m=1}^M \N\big(\z_m|\0, \tau\I\big)$, and a weight vector $\g$ from $p(\g)= \N(\g|\0, \frac{1}{M}\I)$. We then represent the function $f(\cdot)$ in \eqref{eq:value-prob} by 
\begin{align}
	f(\x_{\bi_n}) = \bphi(\x_{\bi_n})^\top \g, \label{eq:rff}
\end{align}
where the input $\x_{\bi_n}$ includes the factors of all the nodes associated with $\bi_n$. Since the sociabilities are often very small and close to zero, we use their softmax parameters $\{\ttomega^k_{i_{nk}}\}$ in $\x_{\bi_n}$. 
When we marginalize out the weight vector $\g$, the joint Gaussian distribution is recovered, and the kernel takes the form of the Monte-Carlo approximation. We will keep $\g$ to prevent computing the full covariance matrix, and to scale to a large number of observed entries.

Combining \eqref{eq:beta-dist}\eqref{eq:dir-dist}\eqref{eq:edge-prob}\eqref{eq:rff}\eqref{eq:value-prob},  we obtain the joint probability of our model, 
\begin{align}
	&p(\text{Joint}) = \prod\nolimits_{k=1}^K p(\bbeta^k) \left(\prod\nolimits_{j=1}^{D_k} \text{Uniform}(\btheta^k_j|[0, \alpha]^{R_1})\right) \notag \\
	&\cdot \left(\prod\nolimits_{r=1}^{R_2} \text{Gamma}(\gamma^k_r|1, \talpha)  \text{Dir}(\bomega^k_r|\gamma^k_r\bbeta^k)\right) \N(\g|\0, \frac{1}{M}\I)\notag \\
	&\cdot  \prod\nolimits_{m=1}^M \N(\z_m|\0, \tau\I)  \prod\nolimits_{n=1}^N w_{\bi_n} \cdot \N(y_{\bi_n}|\bphi(\x_{\bi_n})^\top \g, \sigma^2). \notag 
\end{align}

We use variational inference~\citep{wainwright2008graphical} for model estimation. We introduce a variational posterior $q(\g) = \N(\g|\bmu, \L\L^\top)$, and construct a variational evidence lower bound (ELBO), $\Lcal = \EE_q\left[{\log p(\text{Joint})} - \log{q(\g)}\right]$. We maximize the ELBO to estimate $q(\g)$ and all the other parameters,  including $\{\tbeta^k\}$, $\{\btheta_j^k\}$, $\{\gamma_{r}^k, \tomega^k_r\}$, $\Z$, \etc  Due to the additive structure over the observed entries in $\Lcal$, we can use mini-batch stochastic optimization for efficient and scalable estimation. 


\noindent \textbf{Complexity.} The time complexity of our inference is $O(NM^2+R\sum_{k=1}^K D_k)$ where $R= R_1 + R_2$, and  $N$ is the number of observed entries. Since $M$ is fixed and $M\ll N$, the time complexity is linear in $N$. The space complexity  is $O(M^2+ M +\sum_kD_kR)$, which is to store the latent factors, the frequencies, and the variational posterior.

\section{Related Work}
Numerous tensor factorization methods have been developed, \eg ~\citep{ShashuaH05,Chu09ptucker,sutskever2009modelling,acar2011scalable,hoff_2011_csda,kang2012gigatensor,YangDunson13Tensor,RaiDunson2014,choi2014dfacto,hu2015zero,zhao2015bayesian,raiscalable,du2018probabilistic}. While most methods adopt the multilinear forms of the classical  Tucker~\citep{Tucker66} and CP decompositions~\citep{Harshman70parafac},   a few GP based factorization models~\citep{xu2012infinite,zhe2015scalable,zhe2016dintucker,zhe2016distributed,pan2020streaming} were recently  proposed to capture the nonlinear relationships between the tensor nodes. 

Most methods model dense tensors. From the Bayesian viewpoint, they  all belong to random function prior models~\citep{LloydOGR12randomgraph} on exchangeable arrays. Recently, \citet{tillinghast2021nonparametric} used Gamma processes (\gaps), a commonly used completely random measure (CRM)~\citep{kingman1967completely,kingman1992poisson,lijoi2010models}, to construct a Poisson random measure that represents asymptotically sparse tensors. Their work can be viewed as an extension of the pioneer works~\citet{caron2014sparse,williamson2016nonparametric,caron2017sparse} in sparse random graph modeling. However, their single-level \gaps and accordingly single-level Dirichlet processes (DPs) for finite projection impose a severe restriction on the type of the learned factors. For each tensor node, either all the factors are exclusively sociability factors or only one factor can be the sociability factor. The former misses the location factors that represent the intrinsic properties of the node, and the latter is often inadequate to capture the extrovert interaction activity across different communities. To overcome this limitation, our model uses hierarchical Gamma processes (H\gaps) to construct the sparse tensor model. Due to the sharing of the locations of \gaps in the second level, our model does not have the node alignment issue and is able to estimate an arbitrary set of sociability and location factors to fully capture the extrovert activities and inward attributes respectively, hence giving more interpretability and an additional flexibility to select the number of two types of factors. We couple the normalized H\gaps and GPs to sample both the sparse entry indices (hyperlinks) and entry values, so as to absorb both the sparse structural knowledge and complex relationships into the factor estimates. We point out that the normalized H\gap is HDP~\citep{teh2006hierarchical}, an important extension of DP~\citep{ferguson1973bayesian}. HDPs are popular in nonparametric mixture modeling in text mining~\citep{hong2010empirical,wang2011online,mcfarland2013differentiating},  which is often referred to as topic models. For inference, \citet{tillinghast2021nonparametric} directly used the stick-breaking construction of DPs, which, however, are not available for our model, because it is intractable for HDP inference. Instead, we used DP measures over a finite partition of the sample space to integrate the measures of non-active nodes, and density transformation to obtain the density of finite point masses. The strategy was also used in HDP mixture models, \eg ~\citep{liang2007infinite,bryant2012truly}. \citet{schein2016bayesian}  used Gamma and Poisson distributions to develop a Bayesian Poisson Tucker decomposition model. They discussed that if one uses a Gamma process instead of the Gamma distribution to sample the latent core tensor, the  core tensor will be asymptotically sparse.  However, it is unclear if the tensor itself will also be sparse. 

Another recent line of research~\citep{zhe2018stochastic,pan2020scalable,wang2020self} uses GPs to construct different point processes for temporal events modeling and embedding the participants of these events.  While these problems are formulated as decomposing a special type of tensors, \ie event-tensors, where each entry is a sequence of temporal events (rather than numerical values), these models focus on building expressive non-Poisson processes to capture complex temporal dependencies between the events, \eg the Hawkes processes with local or global triggering effects~\citep{zhe2018stochastic,pan2020scalable}, the non-Poisson, non-Hawkes process~\citep{wang2020self} that captures the long-term, short-term, triggering and inhibition effects.

\section{Experiment}
\vspace{-0.1in}
\subsection{Simulation}
\begin{figure*}
	\centering
	\setlength{\tabcolsep}{0pt}
	\captionsetup[subfigure]{aboveskip=1pt,belowskip=0pt}
	\begin{tabular}[c]{ccccc}
		\setcounter{subfigure}{0}
		\begin{subfigure}[t]{0.2\textwidth}
			\centering
			\includegraphics[width=\textwidth]{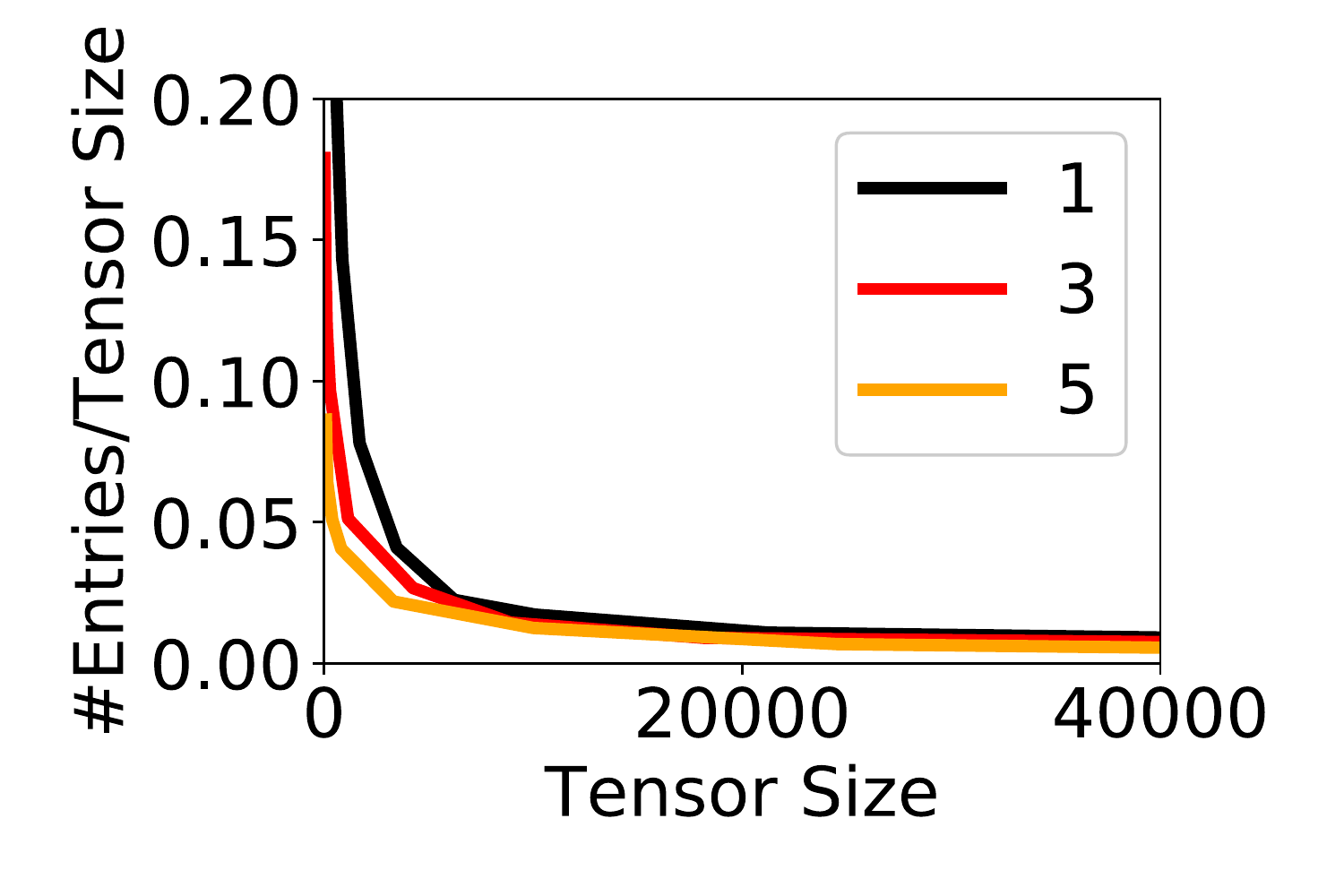}
			\caption{\small Sparsity: Ours}
		\end{subfigure}
	    &
	    	\begin{subfigure}[t]{0.2\textwidth}
	    	\centering
	    	\includegraphics[width=\textwidth]{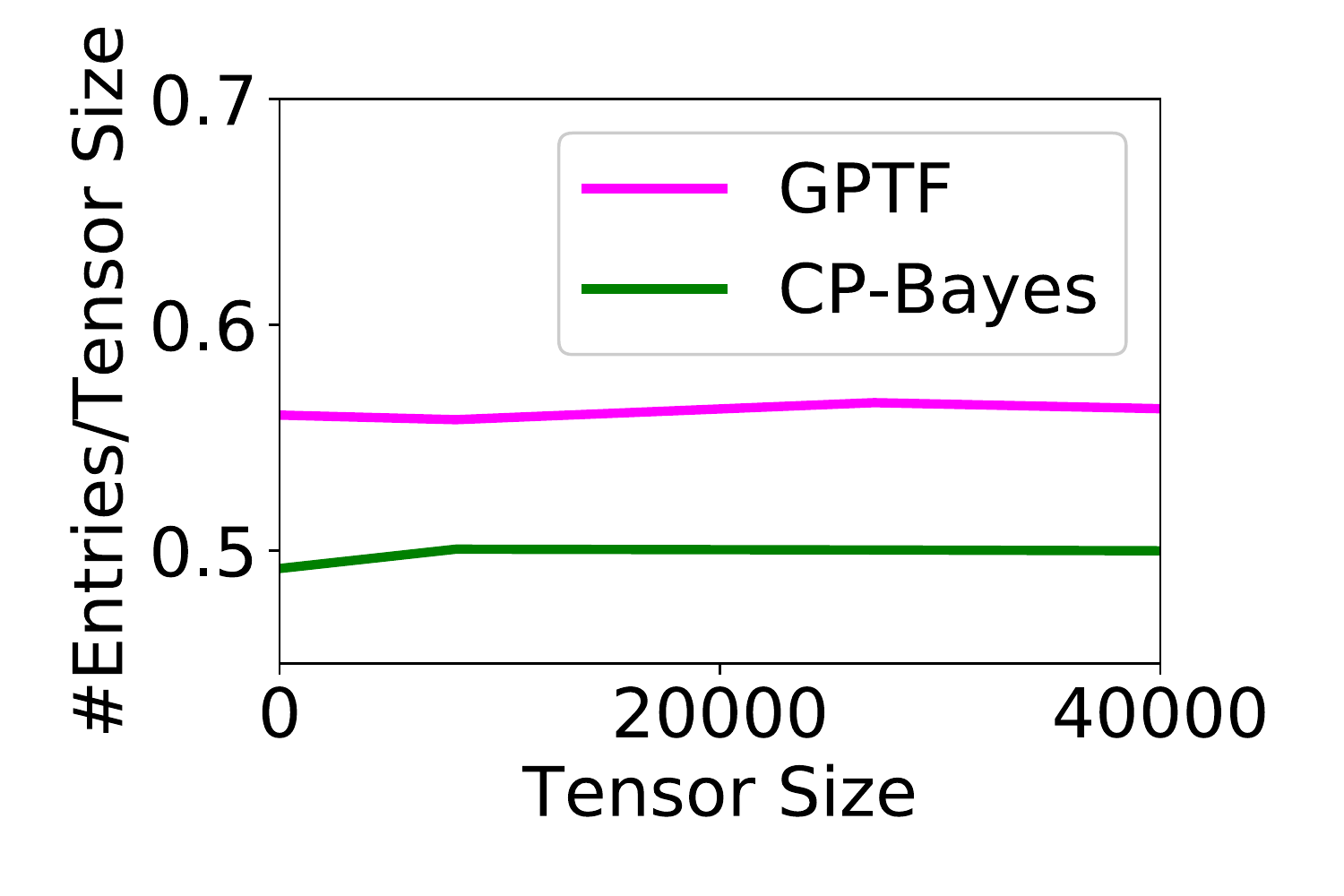}
	    	\caption{\small Sparsity: Dense models}
	    \end{subfigure}
    	&
		\begin{subfigure}[t]{0.2\textwidth}
			\centering
			\includegraphics[width=\textwidth]{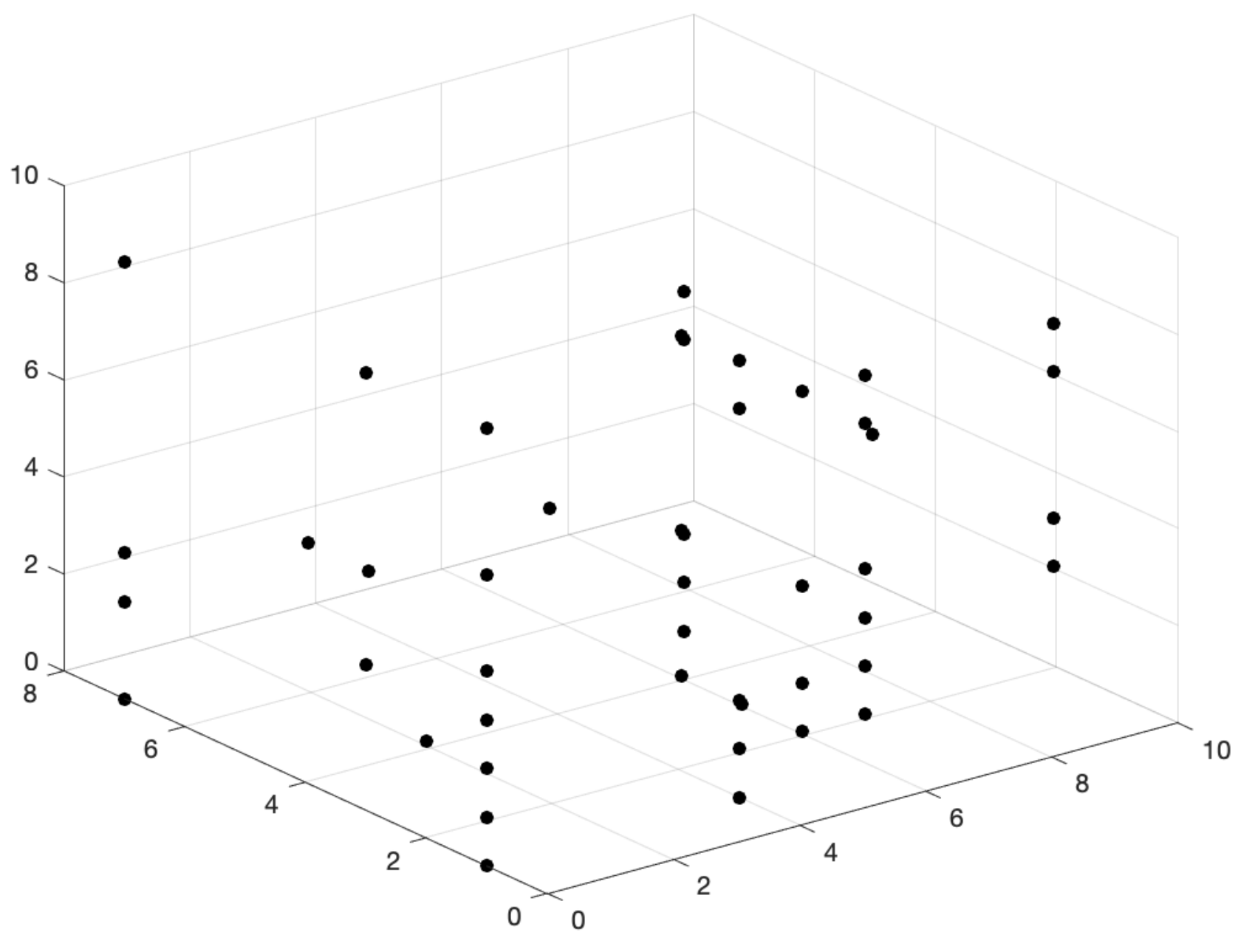}
			\caption{\small Our model}
		\end{subfigure}
		&
		\begin{subfigure}[t]{0.2\textwidth}
			\centering
			\includegraphics[width=\textwidth]{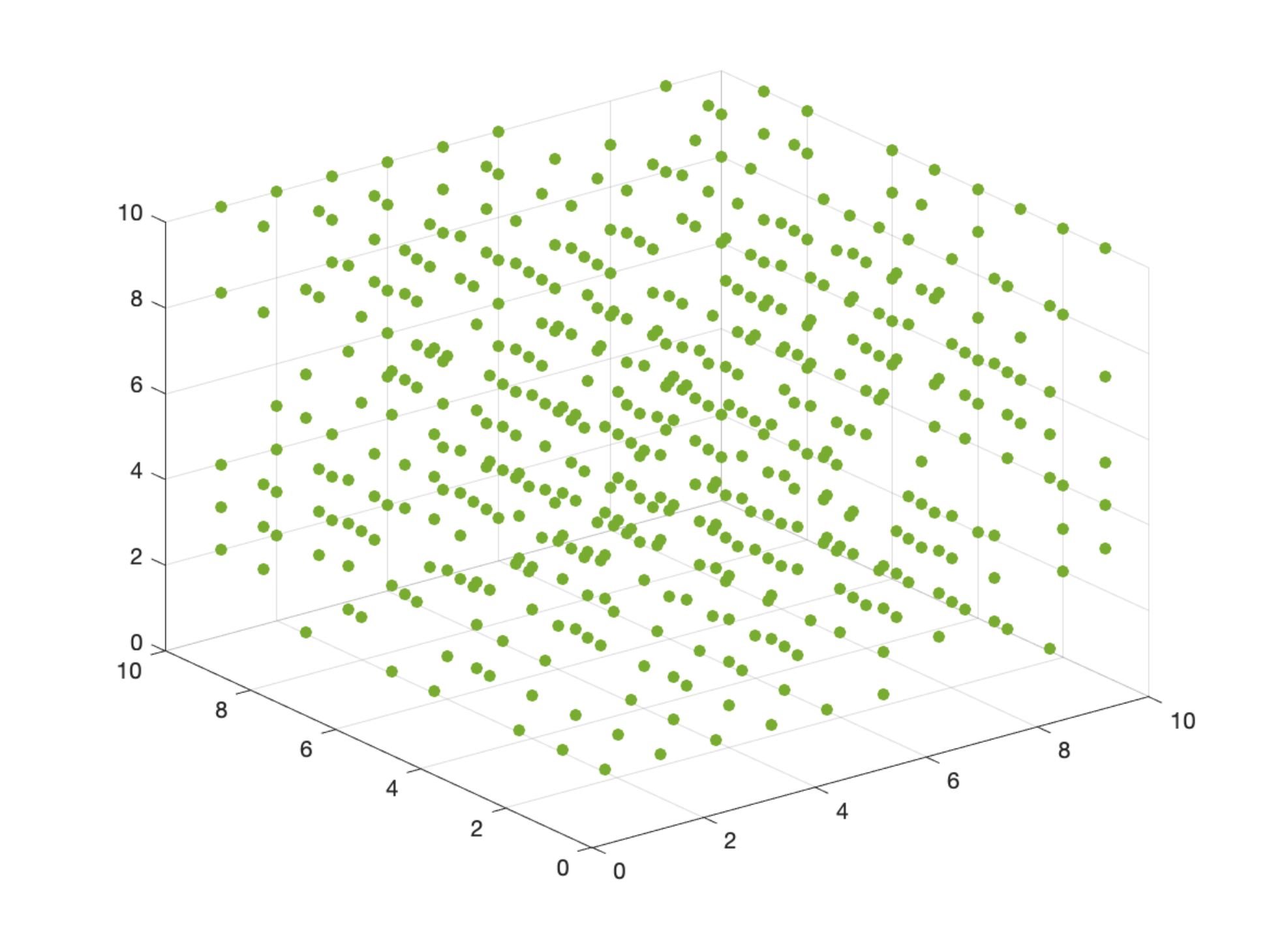}
			\caption{\small CP-Bayes}
		\end{subfigure}
		&\begin{subfigure}[t]{0.2\textwidth}
			\centering
			\includegraphics[width=\textwidth]{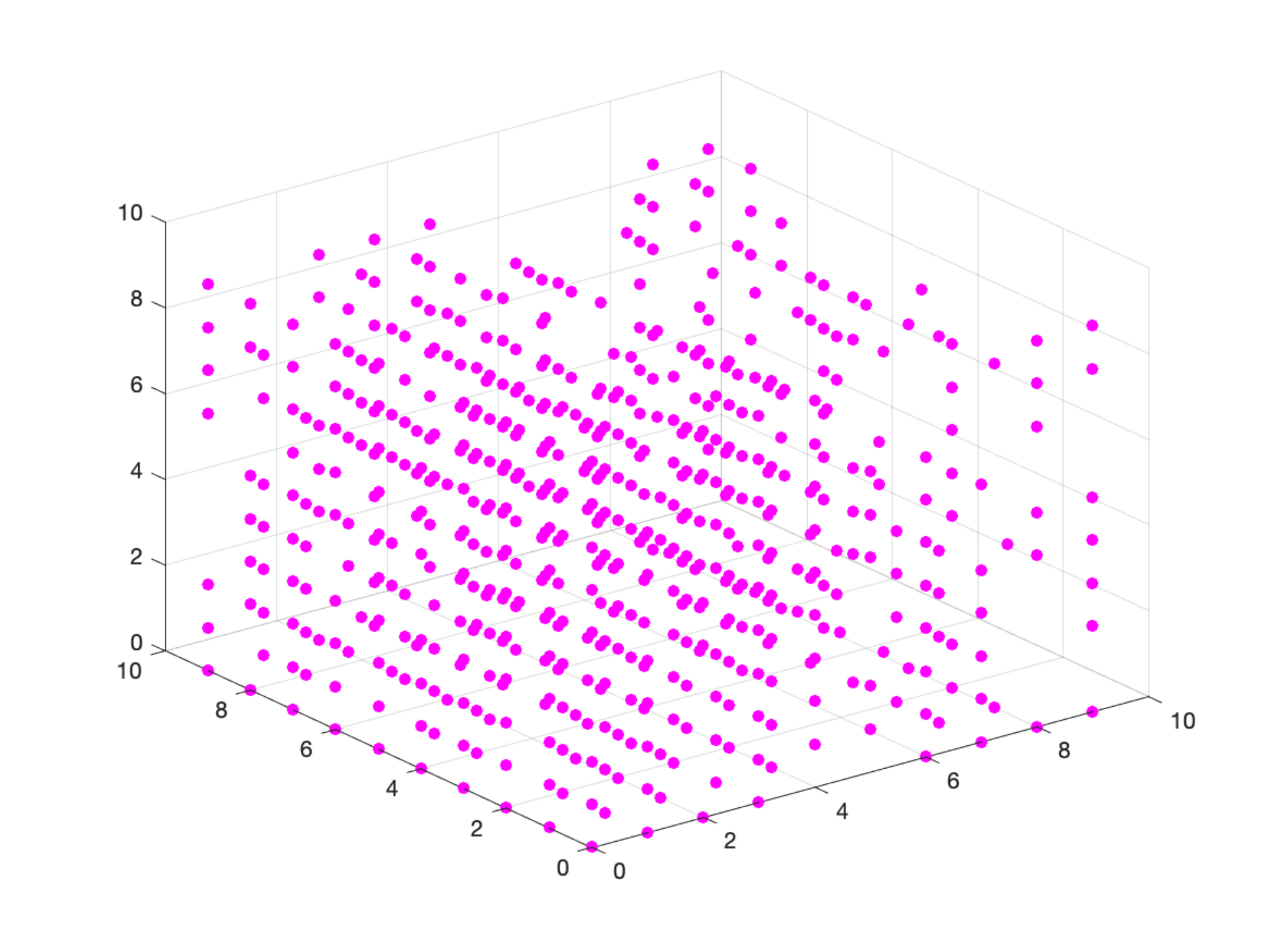}
			\caption{\small GPTF}
		\end{subfigure}
	\end{tabular}
	\vspace{-0.1in}
	\caption{\small The ratio between the sampled entries and tensor size (a, b), and   exemplar tensors generated by each model (c, d, e), of size $10\times 10 \times 10$. The legend in (a) indicates the number of sociability factors. \cmt{Dots in (c, d, e) represent the sampled entries.}} \label{fig:simulation}
	\vspace{-0.1in}
\end{figure*}

\vspace{-0.1in}
First, we checked if our H\gap based tensor process can indeed produce sparse tensors. We used our model to generate a series of tensors with increasingly more present entries, and examined how the proportion of the present entries varied accordingly. The technique details about sampling is given in the Appendix. \cmt{Specifically, we increased $\alpha$ each time, and sampled the total mass of the mean measure of $T$ in \eqref{eq:ppp}, \ie $A([0, \alpha]^{R_1})$. With the total mass as the mean, we sampled the number of present (existent) entries with a Poisson distribution. \cmt{Obviously, the larger $\alpha$, the larger the total mass  and hence the larger the entry number $N$ (with a high probability).} Given the entry number, we used the HDP in \eqref{eq:hdp} to sample the entry indices via the stick-breaking construction (see more details in the Appendix). }We fixed $R_1 = 1$ and varied the number of sociability factors ($R_2$) from $\{1, 3, 5\}$, and $\alpha$ from $[1, 15]$ .\cmt{for $R_2=3$, and $[1, 8]$ for $R_2 = 5$ and $R_2=7$).} Then we looked at the active nodes in the sampled entries (the nodes present in the entry indices), with which we can construct an active tensor that only includes the active nodes in each mode. We calculated the ratio between the number of sampled entries and the size of the active tensor. The ratio indicates the sparsity. For each particular $\alpha$, we ran the simulation for $100$ times and calculated the average ratio, which gives a reliable estimate of the sparsity. We show how the ratio varied along with the size of the active tensor in Fig. \ref{fig:simulation} a. As we can see, the proportion of the sampled entries drops rapidly with the growth of the active tensor and is tending to zero (the gap between the curves and the horizontal axis is decreasing). This is consistent with the (asymptotic) sparse guarantee  in Lemma \ref{lem:2}.  

We contrasted our model with two popular dense factorization models,  CP-Bayes~\citep{zhao2015bayesian,du2018probabilistic} and GPTF~\citep{zhe2016distributed,pan2020streaming},  where we first sampled the latent factors  $\Ucal$ from the standard Gaussian prior distribution, and then independently sampled the presence of each entry $\bi$ from $p(z_\bi|\Ucal) = \text{Bern}\big(z_\bi|f(\x_i)\big)$, where $\x_i = [\u_{i_1}^1; \ldots ; \u_{i_K}^K]$, and $f(\cdot)$ is the factorization function. For CP-Bayes, $f(\cdot)$ is the element-wise version of the CP form, \cmt{$f(\x_{\bi})= \sum\nolimits_{k=1}^K    \1^\top (\u_{i_1}^1  \circ \ldots \circ \u_{i_K}^K)$.} while  for GPTF, $f(\cdot)$ is assigned a GP prior, and we used a sparse approximation with random Fourier feature (the same as in our model inference) to sample a large number of entries. \cmt{We set the frequency number $M = 100$.} We gradually increased the number of nodes in each  mode so as to grow the tensor size. We examined the proportion of the present entries accordingly (\ie $z_\bi = 1$). At each step, we ran the sampling procedure for $100$ times, and computed the average proportion. As shown in Fig. \ref{fig:simulation}b, CP-Bayes and GPTF both generated dense tensors, where the proportion is almost constant, reflecting that the present entry number grows linearly with the tensor size.  In Fig. \ref{fig:simulation}c-e, we showcase exemplar tensors generated by each model. We can see  that the tensors entries sampled from CP-Bayes and GPTF (Fig. \ref{fig:simulation}d and e) are much denser than ours (Fig. \ref{fig:simulation}c)  and more uniform. This is consistent with the fact that their priors are exchangeable and symmetric, and each entry is sampled with the same probability (see Sec. \ref{sect:bk}).

 \begin{figure*}[ht]
 	\vspace{-0.05in}
 	\centering
 	\setlength{\tabcolsep}{0pt}
 	\captionsetup[subfigure]{aboveskip=0pt,belowskip=0pt}
 	\begin{tabular}[c]{ccccc}
 		\setcounter{subfigure}{0}
 		\raisebox{0.1in}
 		{
 			\begin{subfigure}[t]{0.10\textwidth}
 				\centering
 				\includegraphics[width=\textwidth]{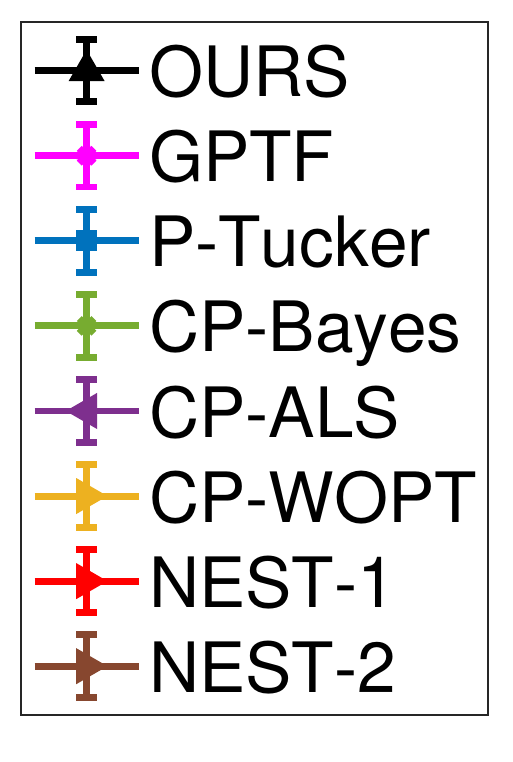}
 			\end{subfigure}
 		}
 		&
 		\begin{subfigure}[t]{0.23\textwidth}
 			\centering
 			\includegraphics[width=\textwidth]{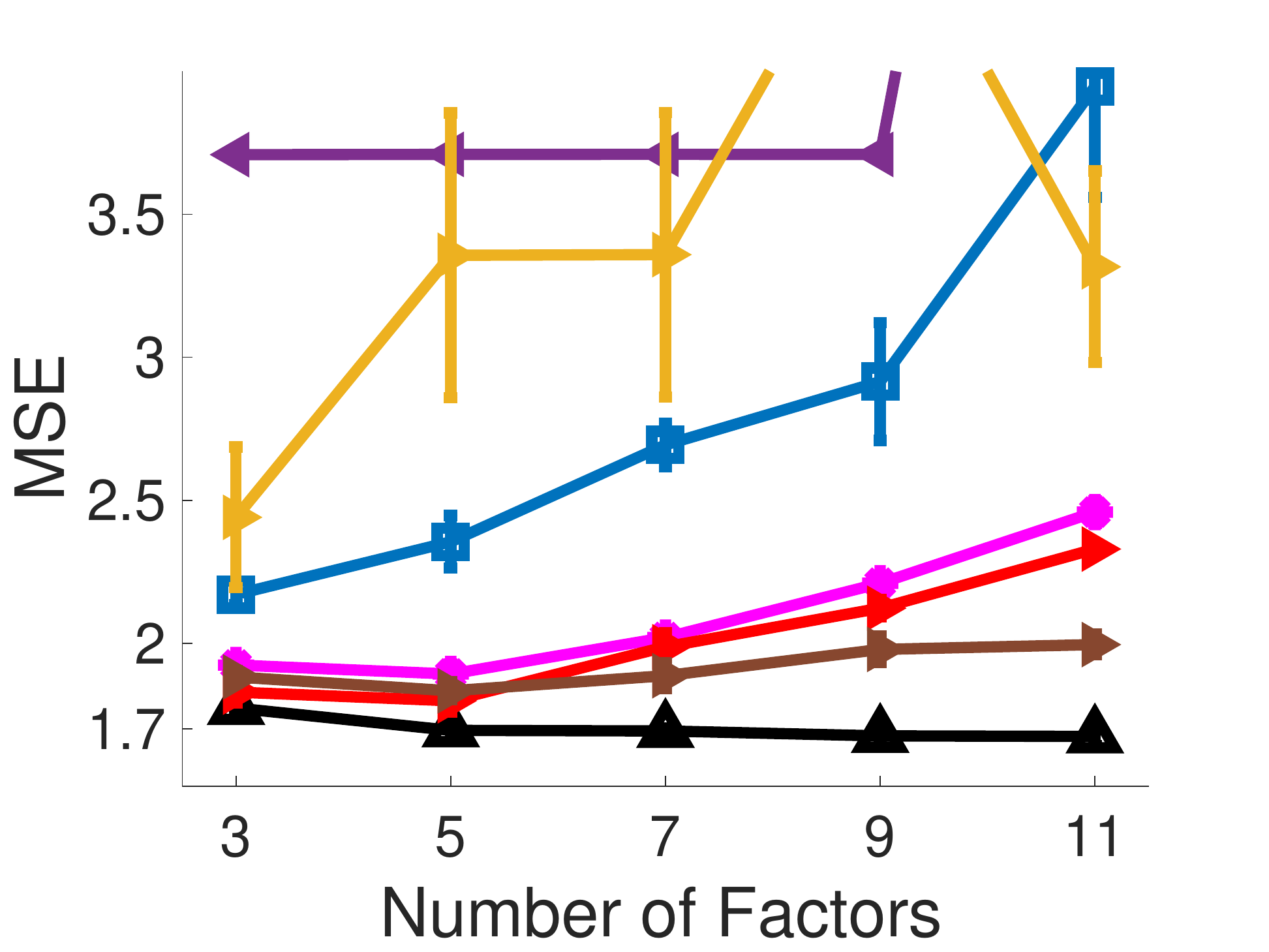}
 			\caption{\small \textit{Alog}}
 		\end{subfigure}
 		&
 		\begin{subfigure}[t]{0.23\textwidth}
 			\centering
 			\includegraphics[width=\textwidth]{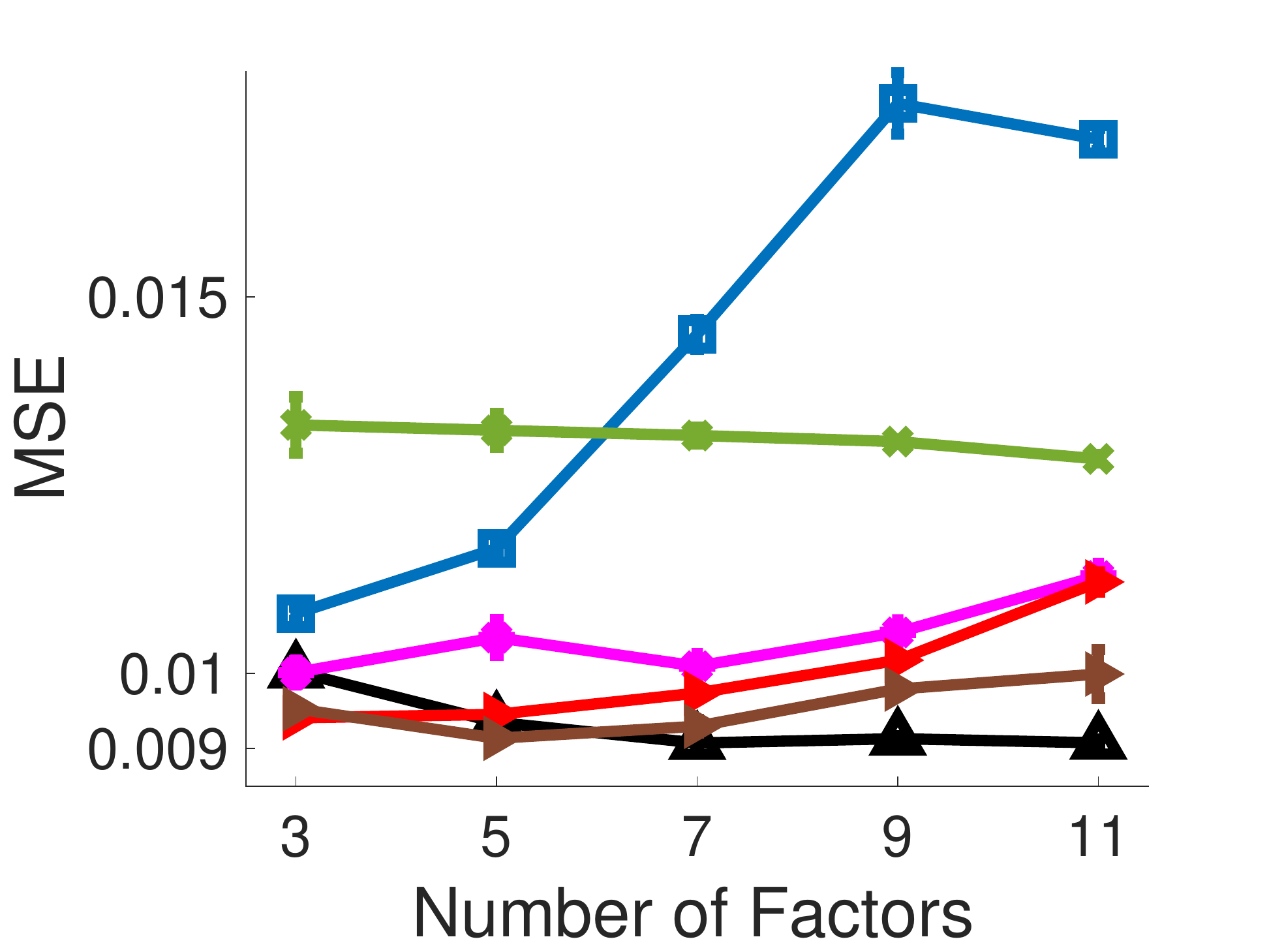}
 			\caption{\small \textit{MovieLens}}
 		\end{subfigure}
 		&
 		\begin{subfigure}[t]{0.23\textwidth}
 			\centering
 			\includegraphics[width=\textwidth]{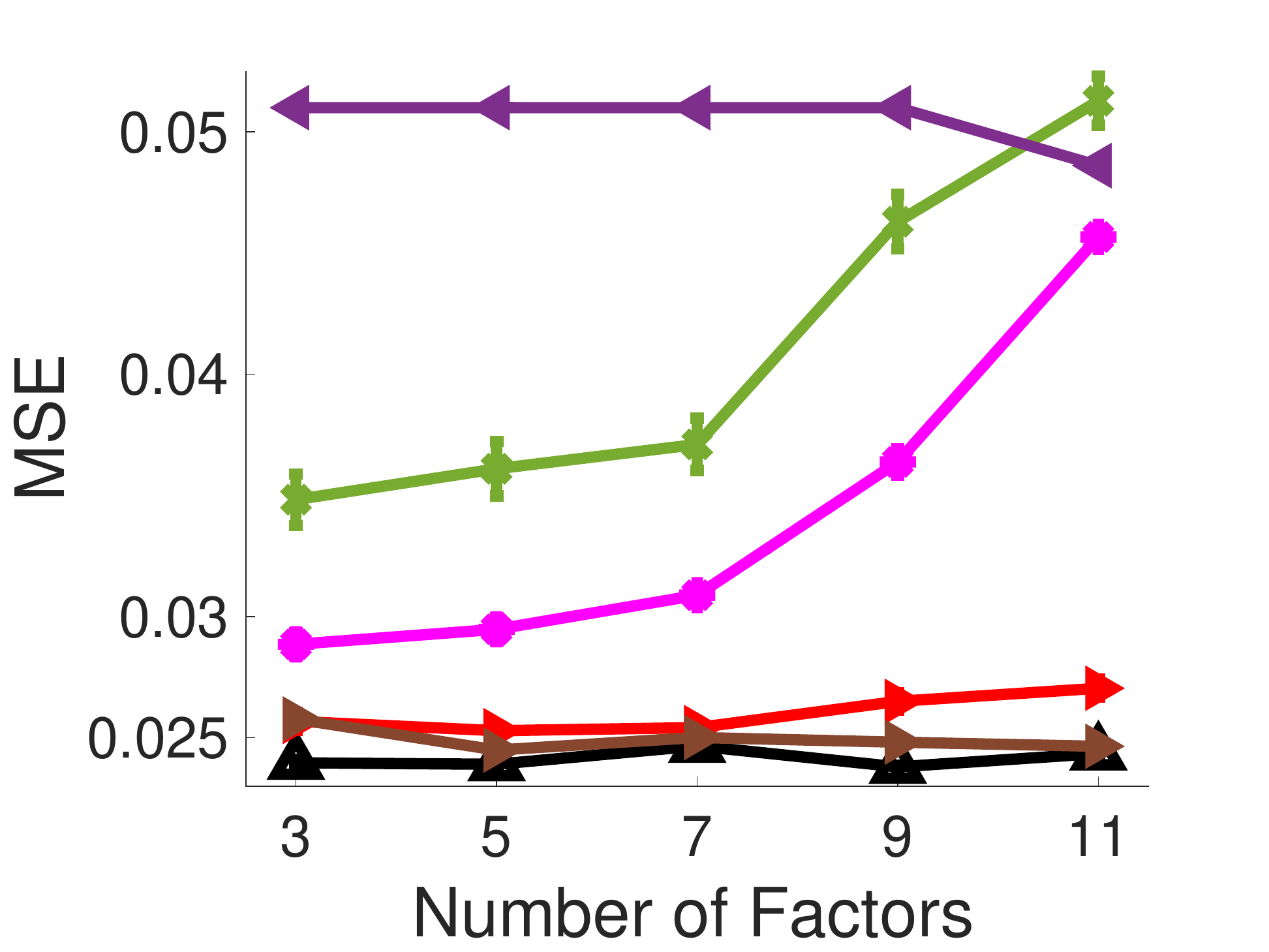}		
 			\caption{\small \textit{SG}}	
 		\end{subfigure}
 		&
 		\begin{subfigure}[t]{0.23\textwidth}
 			\centering
 			\includegraphics[width=\textwidth]{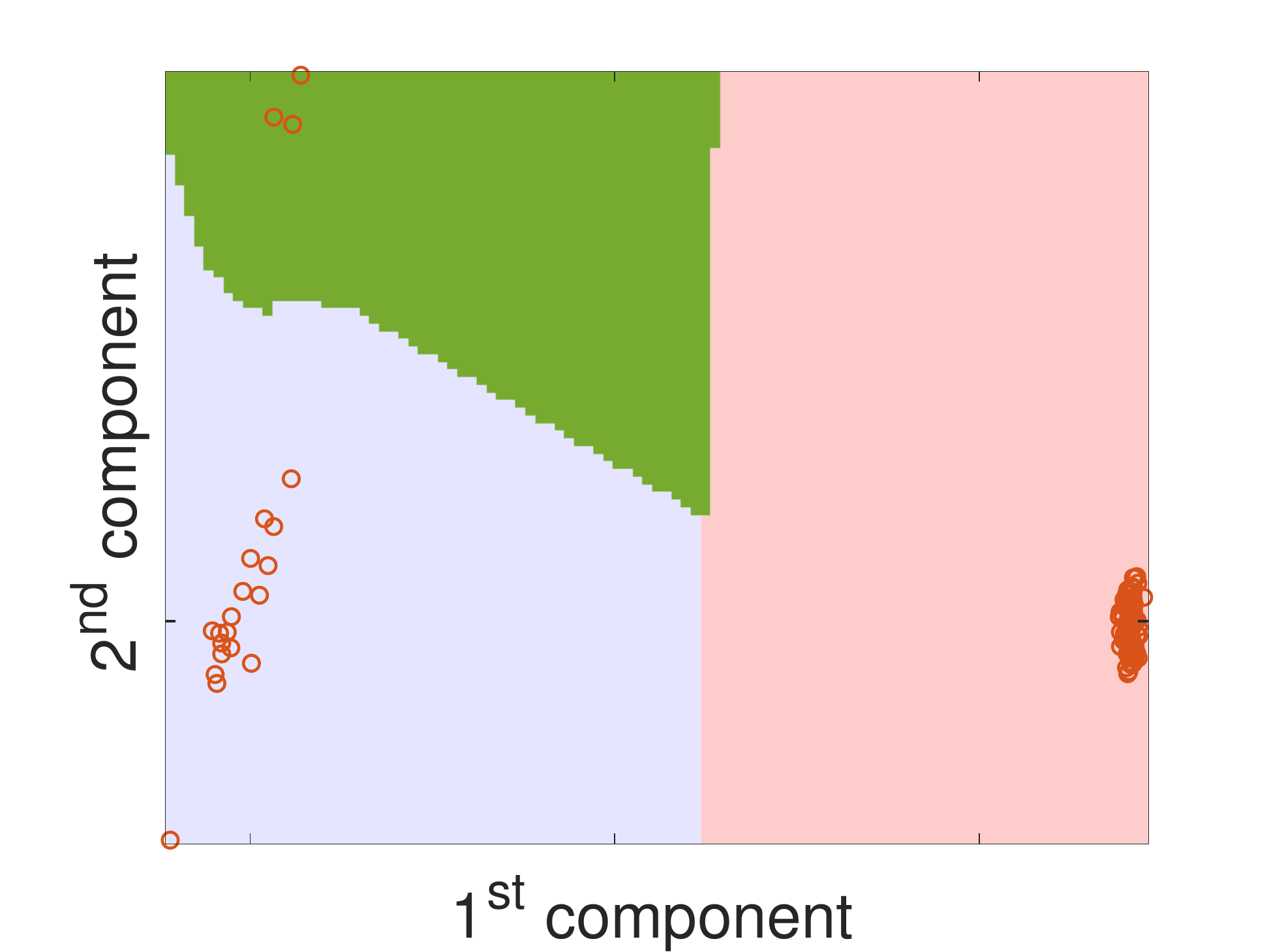}
 			\caption{\textit{Alog}-mode 2}
 		\end{subfigure}
 		\\
 		&
 		\begin{subfigure}[t]{0.23\textwidth}
 			\centering
 			\includegraphics[width=\textwidth]{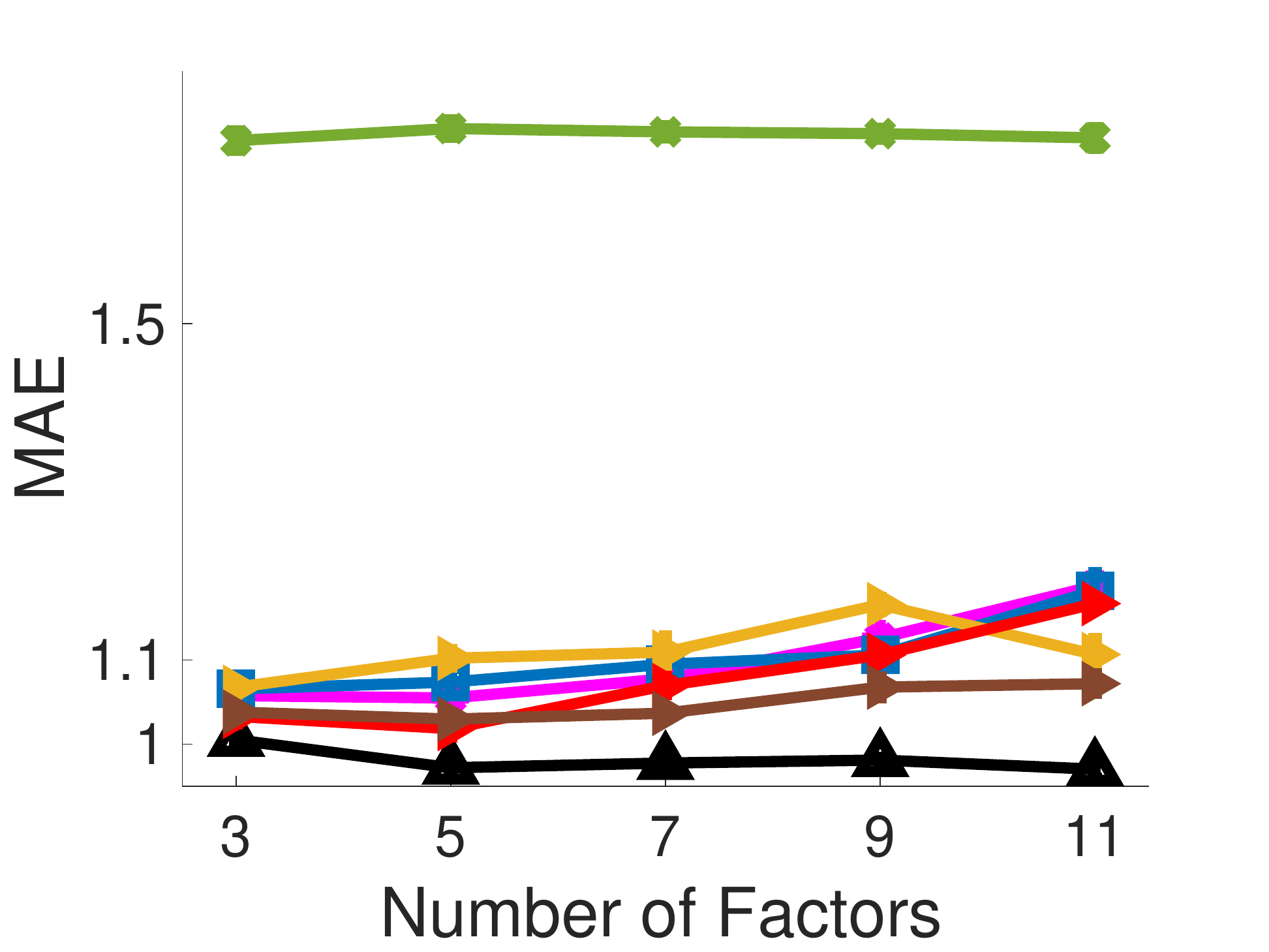}
 			\caption{\small \textit{Alog}}
 		\end{subfigure}
 		&
 		\begin{subfigure}[t]{0.23\textwidth}
 			\centering
 			\includegraphics[width=\textwidth]{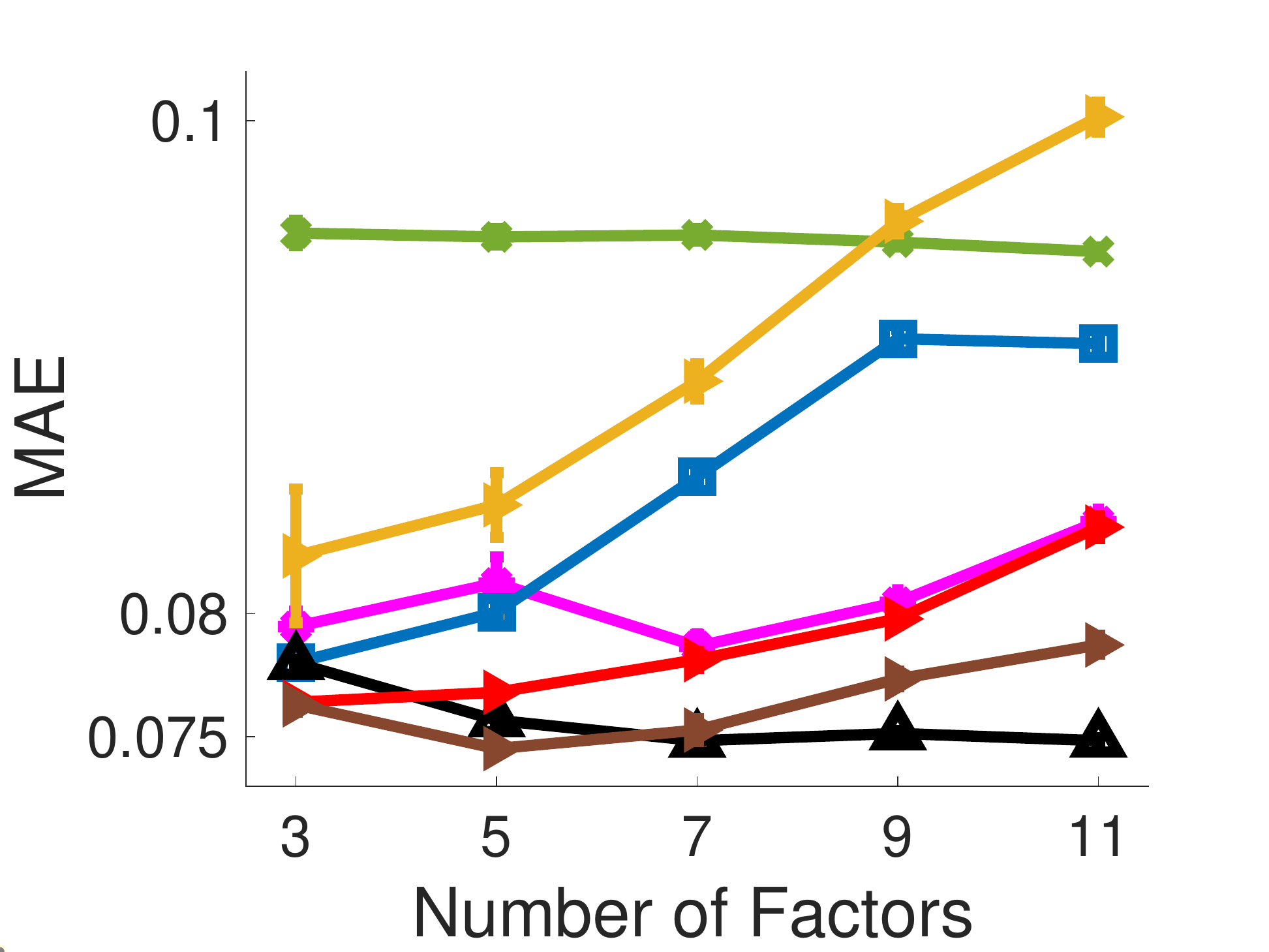}
 			\caption{\small \textit{MovieLens}}
 		\end{subfigure}
 		&
 		\begin{subfigure}[t]{0.23\textwidth}
 			\centering
 			\includegraphics[width=\textwidth]{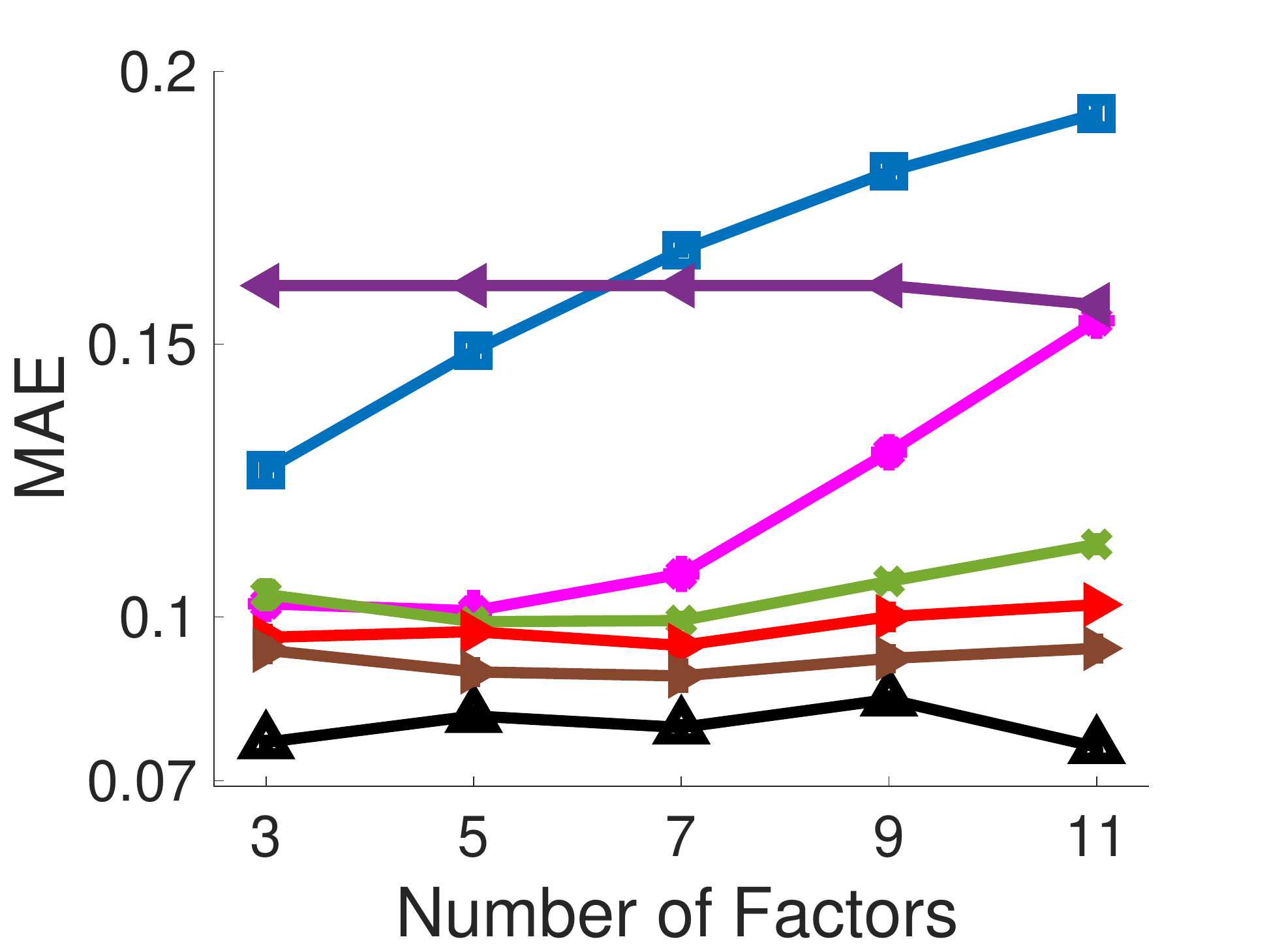}
 			\caption{\small \textit{SG}}
 		\end{subfigure}
 		&
 		\begin{subfigure}[t]{0.23\textwidth}
 			\centering
 			\includegraphics[width=\textwidth]{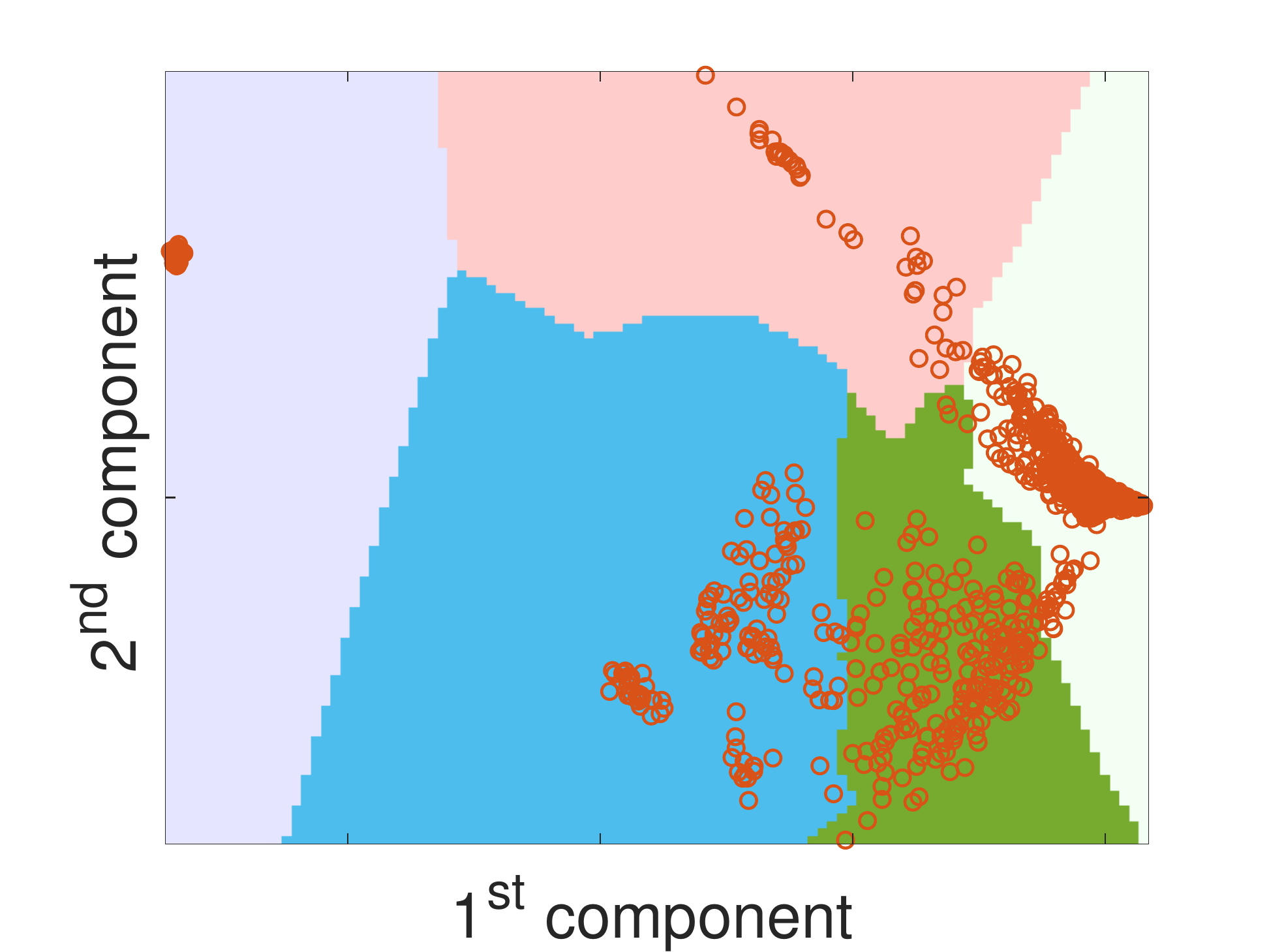}
 			\caption{\textit{MovieLens}-mode 2}
 		\end{subfigure}	 \\
 		&
 		\begin{subfigure}[t]{0.23\textwidth}
 			\centering
 			\includegraphics[width=\textwidth]{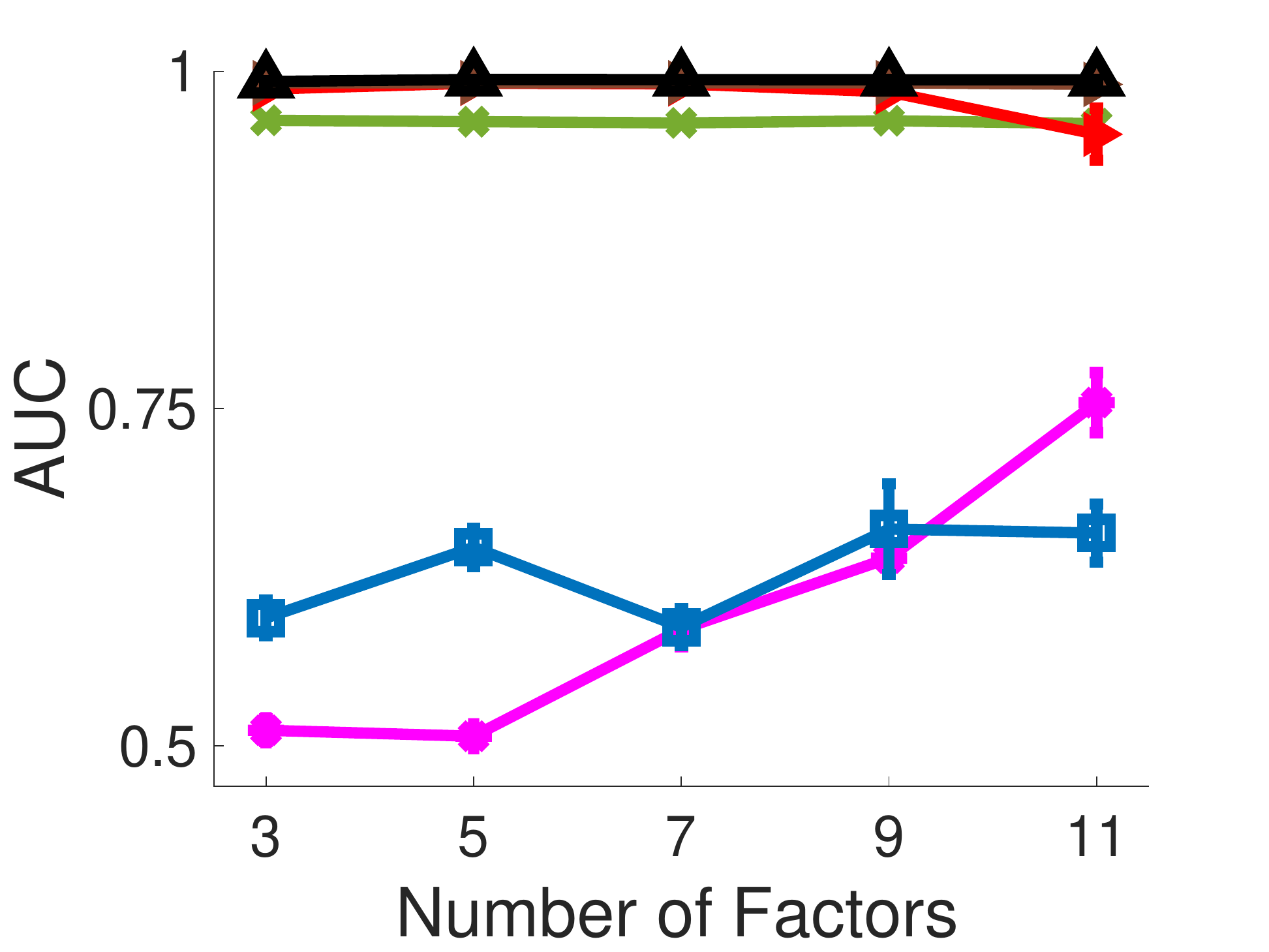}
 			\caption{\small \textit{Alog}-Link}
 		\end{subfigure} 
 		&
 		\begin{subfigure}[t]{0.23\textwidth}
 			\centering
 			\includegraphics[width=\textwidth]{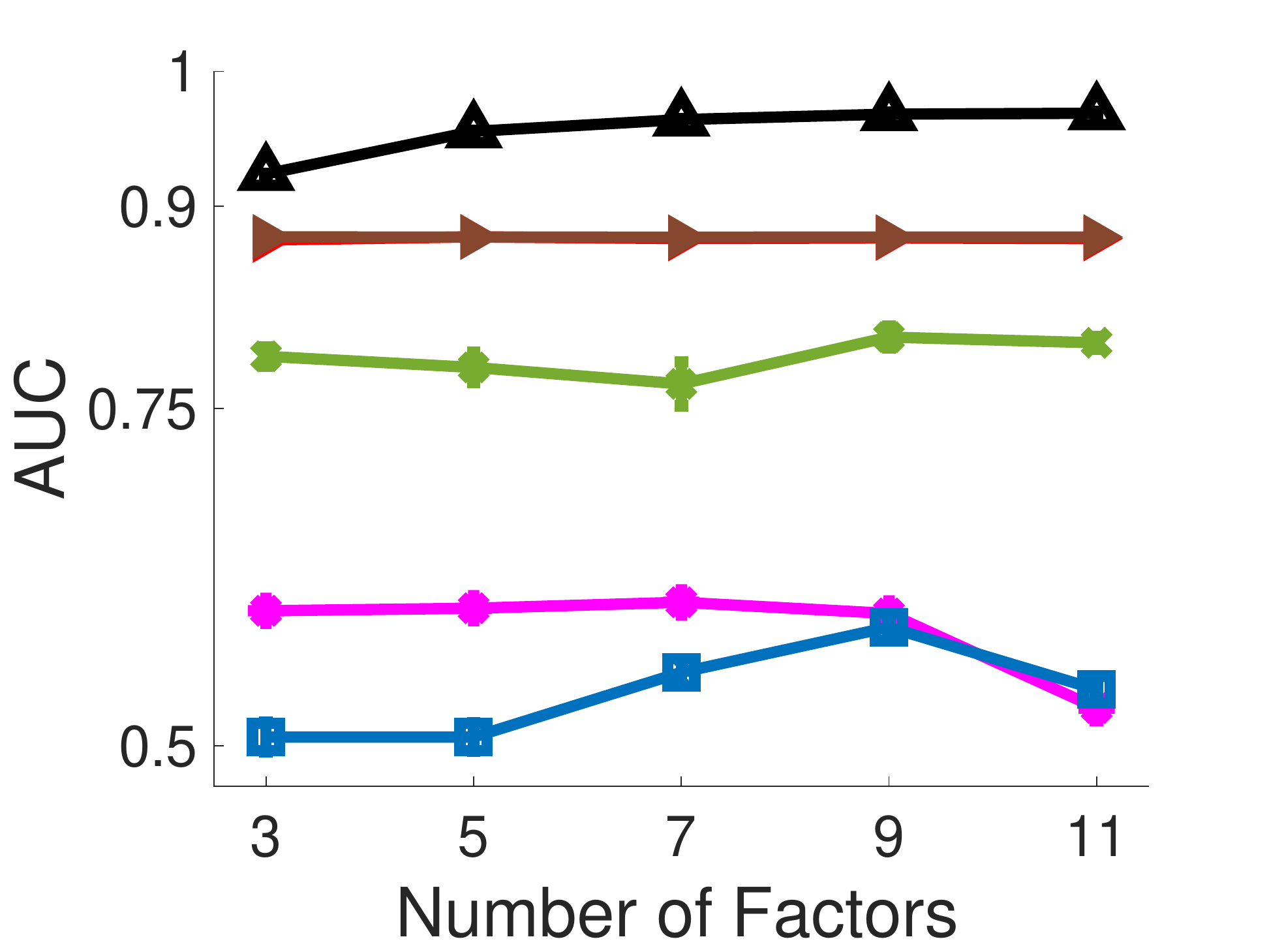}
 			\caption{\small \textit{MovieLens}-Link}
 		\end{subfigure}
 		&
 		\begin{subfigure}[t]{0.23\textwidth}
 			\centering
 			\includegraphics[width=\textwidth]{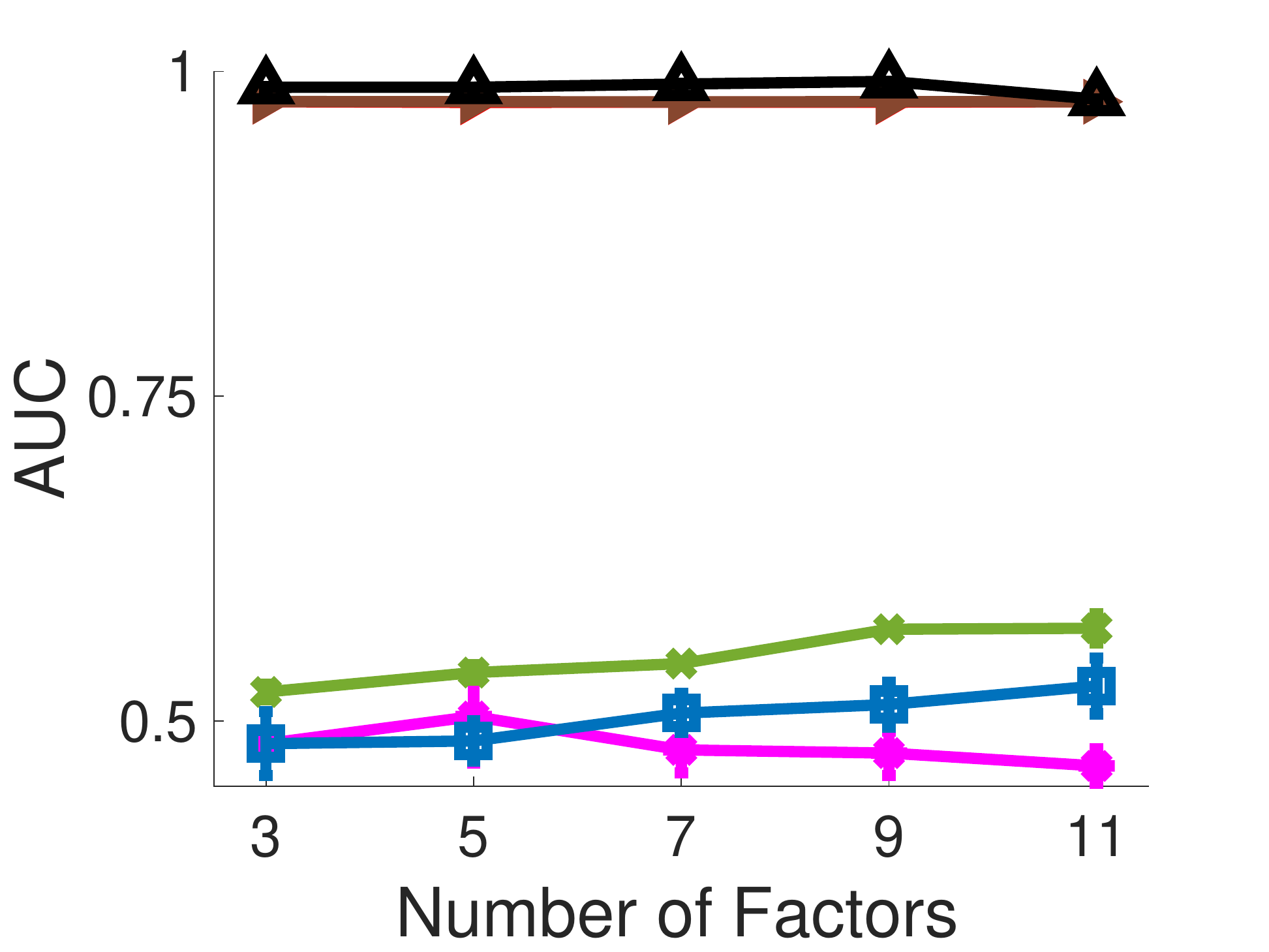}
 			\caption{\small \textit{SG}-Link}
 		\end{subfigure}
 		&
 		\begin{subfigure}[t]{0.23\textwidth}
 			\centering
 			\includegraphics[width=\textwidth]{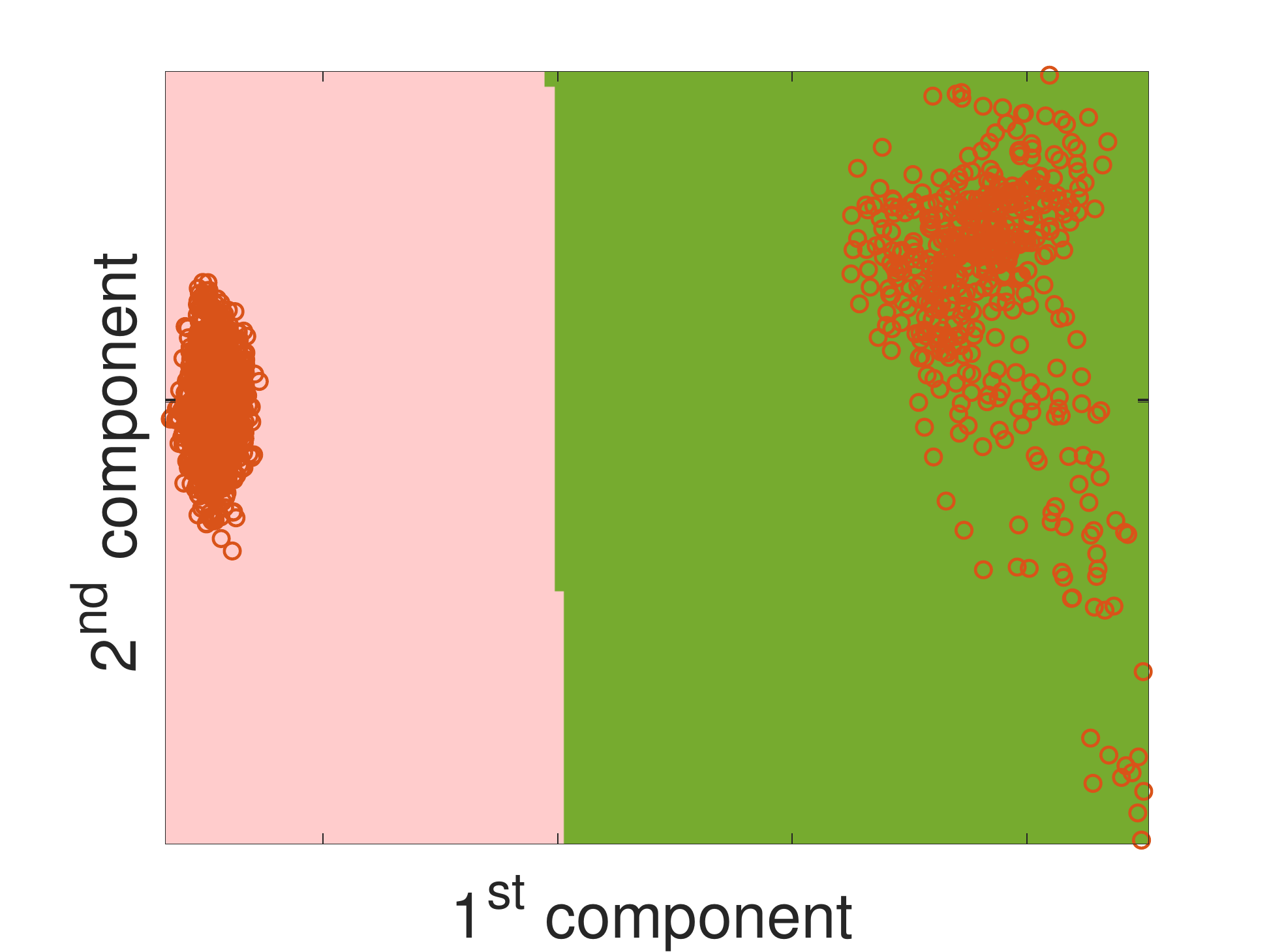}
 			\caption{\textit{SG}-mode 3}
 		\end{subfigure}
 	\end{tabular}
 \vspace{-0.1in}
 	\caption{\small Prediction accuracy of entry values (a-c, e-g) and entry indices (i-k), and the structures of the estimated factors by our method (d, h, l). Note that, the results of some approaches were much worse than the others and were not included, \eg the broken curves of CP-WOPT and CP-ALS in (a); NEST-1 and NEST-2 overlap in (j, k), and NEST-2 overlaps with Ours in (i).} 
 	\label{fig:pred-acc}
 	\vspace{-0.2in}
 \end{figure*}
\vspace{-0.1in}
\subsection{Predictive Performance in Practical Applications}
\vspace{-0.1in}
Next, we evaluated the predictive performance in three real-world benchmark datasets: (1)  \textit{Alog}~\citep{zhe2016distributed}, extracted from a file access log, depicting the access frequency among \textit{users}, \textit{actions}, and \textit{resources}, of size $200 \times 100 \times 200$. The value of each present entry is the logarithm of the access frequency. The proportion of present entries is $0.3\%$. (2)  \textit{MovieLens} (\url{https://grouplens.org/datasets/movielens/100k/}), a three-mode \textit{(user, movie, time)} tensor, of size  $1000 \times 1700 \times 31$. There are 100K present entries, taking $0.19\%$ of the whole tensor. The entry values are movie ratings in $[0, 10]$. We divided every value by $10$ to normalized them in $[0, 1]$. (3) \textit{SG}~\citep{li2015rank,costco},  extracted from Foursquare Singapore data, representing \textit{(user, location, point-of-interest)} check-ins, of size $2321 \times 5596 \times 1600$. The value of each present entry is the normalized check-in frequency in $[0, 1]$. The number of present entries is $105,764$, taking $0.0005\%$ of whole tensor.  

\noindent\textbf{Methods and Settings.} We compared with the following state-of-the-art tensor factorization methods. (1) {CP-Bayes}~\citep{zhao2015bayesian,du2018probabilistic}, a Bayesian CP factorization model with Gaussian likelihood. (2) {CP-ALS~}\citep{TTB_Software},  CP factorization using alternating least squares to update the latent factors. (3)  {CP-WOPT}~\citep{acar2011scalable},  a fast CP factorization method using conjugate gradient descent to optimize the latent factors. (4) P-Tucker~\citep{oh2018scalable}, a highly efficient Tucker factorization algorithm that updates the factor matrices row-wisely in parallel. (5) {GPTF} ~\citep{zhe2016distributed,pan2020streaming}, nonparametric tensor factorization based on GPs. It is the same as our modeling the entry value as a latent function of the latent factors (see \eqref{eq:value-prob}), except that it samples the latent factors from a standard Gaussian prior. The same sparse GP approximation, \ie random Fourier features, was employed for scalable model estimation.  (6) NEST-1 and (7) NEST-2, which are the two models proposed by \citet{tillinghast2021nonparametric}. NEST-1 uses a single DP in each mode to sample the sparse tensors, and can only estimate one sociability factor for each node; the remaining factors are location factors (from DP locations). NEST-2 uses multiple GEM distributions to draw DP weights for sparse tensor modeling, hence all the factors are sociability factors and there are no location factors.  We implemented our method with PyTorch~\citep{paszke2019pytorch}. CP-Bayes, GPTF, NEST-1 and NEST-2 were implemented with TensorFlow~\citep{abadi2016tensorflow}. Since all these methods used stochastic mini-batch optimization, we chose the learning rate from $\{10^{-4}, 2 \times 10^{-4}, 5\times 10^{-4}, 10^{-3}, 5\times 10^{-3}, 10^{-2}\}$, and set the mini-batch size to $200$ for \textit{Alog} and \textit{MovieLens}, and $512$ for \textit{SG}. We ran 700 epochs on all the three datasets, which is enough for convergence. We used the original MATLAB implementation of CP-ALS and CP-WOPT (\url{http://www.tensortoolbox.org/}),  and C++ implementation of P-Tucker (\url{https://github.com/sejoonoh/P-Tucker}),  and their default settings. Note that  CP-ALS needs to fill nonexisting entries with zero-values to manipulate the whole tensor.

\noindent\textbf{Missing Entry Value Completion.} We first tested the accuracy of predicting the missing entry values. To this end, we randomly split the existent entries in each dataset into $80\%$ for training and $20\%$ for test. We then ran each method, and evaluated the Mean Square Error (MSE) and Mean Absolute Error (MAE). We varied the number of factors $R$ from $\{3, 5, 7, 9, 11\}$, and for each setting, we ran the experiment for five times. Since our approach can flexibly estimate the two types of factors, \ie sociability and location factors, to identify an appropriate trade-off, we ran an extra validation in the training set to select $R_1$ and $R_2$ ($R_1 + R_2 = R$) (In the Appendix, we show how the performance of our method varies along with all possible combinations of $R_1$ and $R_2$). Note that while NEST also estimates the two types of factors,  their number choice is strictly fixed: for NEST-1, $R_1 = 1, R_2 = R-1$, and NEST-2, $R_1 = R, R_2 = 0$. We report the average MSE, average MAE and their standard deviations in Fig. \ref{fig:pred-acc} a-c and e-g. As we can see, 
our method nearly always obtains the best performance, except when $R=3, 5$, our method is slightly worse than NEST-1 and NEST-2. In general, our method and NEST outperforms the dense models by a large margin, which confirm the power of the sparse tensor modeling. Our method further improves upon NEST in most cases, especially when $R$ is relatively bigger ($R=7,9, 11$; see Fig. \ref{fig:pred-acc}a,b,e,f and g),  which demonstrates the advantage of our method that can estimate an arbitrary number of sociability and location factors, and allow the selection of their trade-off. 


\noindent\textbf{Missing Entry (Link) Prediction.} Next, we examined our model in predicting missing entry indices.  To this end, from each dataset, we randomly sampled $80\%$ of existent entries, and used their indices for training. Then we used the remaining $20\%$ existent entries plus ten-times nonexistent entries (randomly generated) for test. We did not incorporate all the nonexistent entries for test in order to prevent  them from dominating the result --- they are too many. We compared with CP-Bayes, GPTF, P-Tucker, NEST-1 and NEST-2. { Note that for CP-Bayes and GPTF, we used a Bernoulli distribution to sample the presence (or existence) of an entry, for P-Tucker we used a zero-thresholding function.} We computed the area under ROC curve (AUC) of predictions on the test entries.\cmt{ (note that for P-Tucker, we used the raw score (not through the thresholding function) to compute the AUC to make it as large as possible).}  We repeated the experiment for five times and report the average AUC and its standard deviation in Fig. \ref{fig:pred-acc} i-k. We can see that our method shows much better prediction accuracy than all the dense models. On \textit{Alog} and \textit{SG} dataset, the prediction accuracy of our method is close to or slightly better  than NEST (\ie Fig. \ref{fig:pred-acc}i and g); However, on \textit{MovieLens} dataset, our method significantly outperforms NEST by a large margin. See Fig. \ref{fig:pred-acc}j; note that the results of NEST-1 and NEST-2 overlap.  
It is worth noting that the competing dense methods perform much worse on \textit{SG} than on \textit{Alog} and \textit{MovieLens}. This might because \textit{SG} is much sparser (0.0005\% \textit{vs.} 0.3\% and 0.19\%), and the model misspecification is more severe, leading to much inferior accuracy. 
\vspace{-0.1in}
\subsection{Pattern Discovery}
\vspace{-0.1in}
Finally, we investigated if our method can discover hidden patterns within the tensor nodes, as compared with the popular dense factorization models. To this end, we set $R=11$ where $R_ 1 = 5$ and $R_2 = 6$, and ran our method on all the three datasets. We then used Principal Component Analysis (PCA) to project the learned factors in each mode onto a plain. The positions of the points represent the first and second principal components.  To find the patterns, we ran the k-means algorithm and filled the cluster regions with different colors. We used the elbow method~\citep{ketchen1996application} to select the cluster number. In Fig. \ref{fig:pred-acc}d, h and l, we show the results of the second mode on \textit{Alog} and \textit{MovieLens} and the third mode on \textit{SG}. Each point corresponds to a tensor node. As we can see, these results reflect clear, interesting clustering structures. By contrast, the factors estimated by CP-Bayes and GPTF, as shown in Fig. 2 of the Appendix, do not exhibit patterns and their principal components are distributed like a symmetric Gaussian.   While the datasets have been completely anonymized and we are unable to look into the meaning of the patterns discovered by our method, together this has demonstrated that our model, by capturing the sparse structural information within the entry indices,  can potentially discover more interesting and important knowledge and further enhance the interpretability. 

\vspace{-0.1in}
\section{Conclusion}
\vspace{-0.1in}
We have presented a novel nonparametric tensor factorization method based on hierarchical Gamma processes. Our model not only can sample sparse tensors  to match the nature of data sparsity in practice, but also can estimate an arbitrary number of sociability and location factors  so as to discover interesting and important patterns.

\bibliographystyle{apalike}
\bibliography{SparseHDP}

\section*{Appendix}
\section{Tensor Sampling Details}
In general, to simulate a Poisson random measure (PRM) with the rate measure $\mu(\cdot)$ on $(\Omega, \Bcal)$ where $\Omega$ is the universal set and $\Bcal$ is a $\sigma$-algebra that includes $\Omega$, we conduct two steps. 
\begin{itemize}
	\item  Sample the number of points $N \sim  \text{Poisson}\big(\mu(\Omega)\big)$.
	\item  Sample $N$ points (locations) i.i.d with the normalized rate measure $\frac{\mu}{\mu(\Omega)}$ (which is a probability measure).
\end{itemize}

We use this standard procedure to sample sparse tensors with our model. Given a particular $\alpha$, we first sample the total mass $$A([0, \alpha]^{R_1}) = \frac{1}{R_2}\sum_{r=1}^{R_2} W^\alpha_{1,r}([0, \alpha]^{R_1}) \times \ldots \times W^\alpha_{K,r}([0, \alpha]^{R_1}).$$ To sample the total mass, we need to sample each $W^\alpha_{k,r}([0, \alpha]^{R_1})$ where  $1\le k \le K$ and $1 \le r \le R_2$, which according to the definition of \gaps, follows  $\text{Gamma}(1, L_k^\alpha([0, \alpha]^{R_1}) )$. Recursively, $L_k^\alpha([0, \alpha]^{R_1})$ is sampled from $\text{Gamma}(1, \lambda_\alpha([0, \alpha]^{R_1}))$. We then sample the number of present entries (\ie points) $N$ from a Poisson distribution with the total mass as the mean. Next, we use the normalized H\gaps , namely, HDPs defined in Sec 3.2 of the main paper to sample each entry. To obtain the probability measure over all possible entries, namely, the normalized rate measure in (8) of the main paper, we first sample the weights $\beta^k_j$ in the first level of DPs (see (6) of the main paper). We use the stick-breaking construction. Usually, when $j>2000$, the weights are below the machine precision and automatically truncated to zero, and we do not need to store an infinite number of weights. Following \citep{teh2006hierarchical}, the stick-breaking construction of the  weights in a second-level DP $H^k_r$ is 
\begin{align}
	\nu^k_{rj} \sim \text{Beta}\left(\gamma^k_r \beta^k_j,  \gamma^k_r \left(1 - \sum_{l=1}^j \beta^k_j\right)\right),\; \omega^k_{rj} = \nu^k_{rj}\prod\nolimits_{t=1}^{j-1} (1-\nu^k_{rt})\; (1 \le j \le \infty). \notag
\end{align}
Accordingly, $\nu^k_{rj}$ will be truncated to zero at which $\beta^k_j$ is zero, and so is the weight $\omega^k_{rj}$. Hence, we obtain a finite set of nonzero weights, with which we calculate the probability measure to sample the present entries. 
\begin{figure*}
	\centering
	\setlength{\tabcolsep}{0pt}
	\captionsetup[subfigure]{aboveskip=1pt,belowskip=0pt}
	\begin{tabular}[c]{ccccc}
		\setcounter{subfigure}{0}
		\begin{subfigure}[t]{0.25\textwidth}
			\centering
			\includegraphics[width=\textwidth]{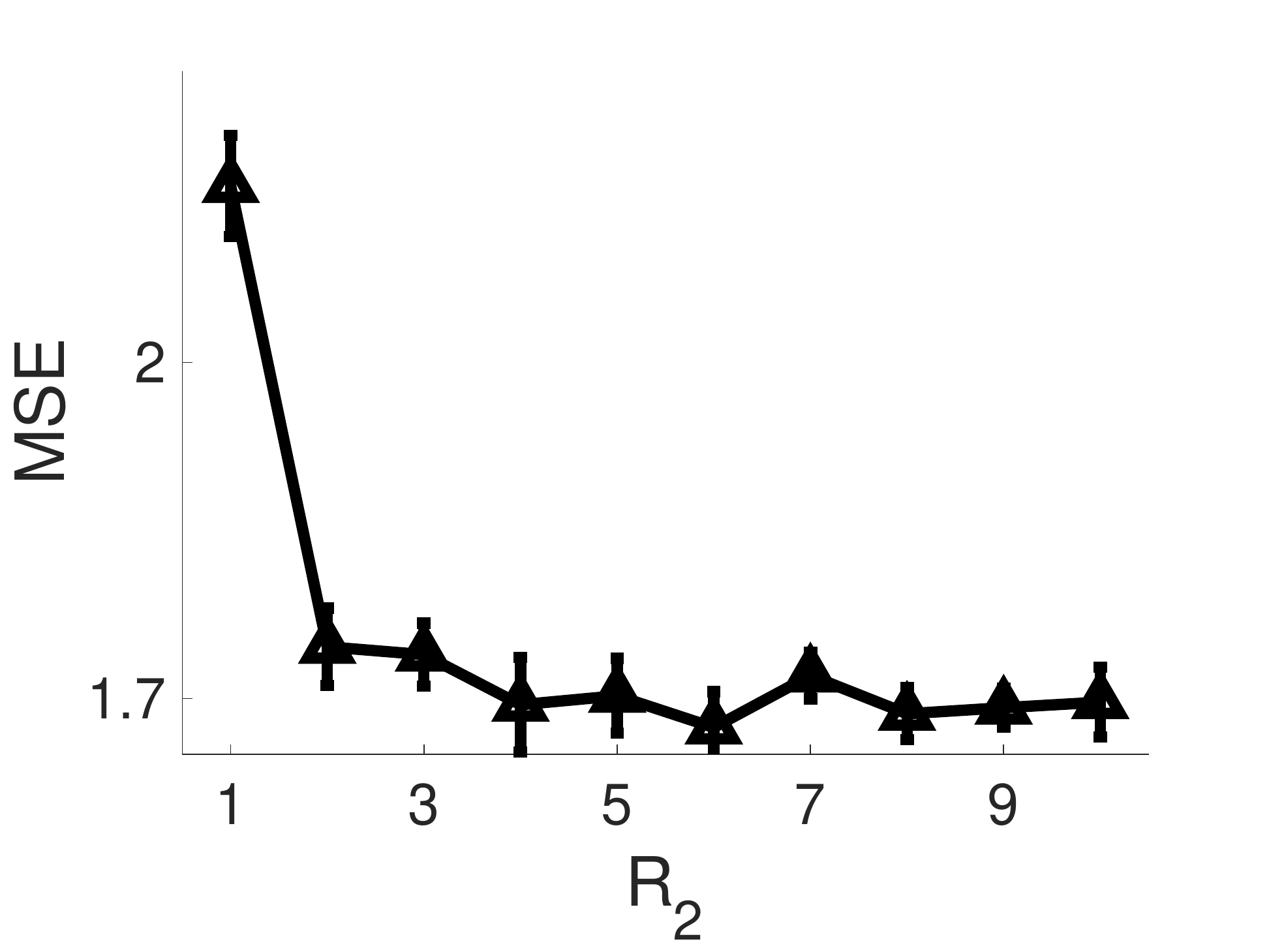}
			\caption{\small \textit{Alog}}
		\end{subfigure}
		&
		\begin{subfigure}[t]{0.25\textwidth}
			\centering
			\includegraphics[width=\textwidth]{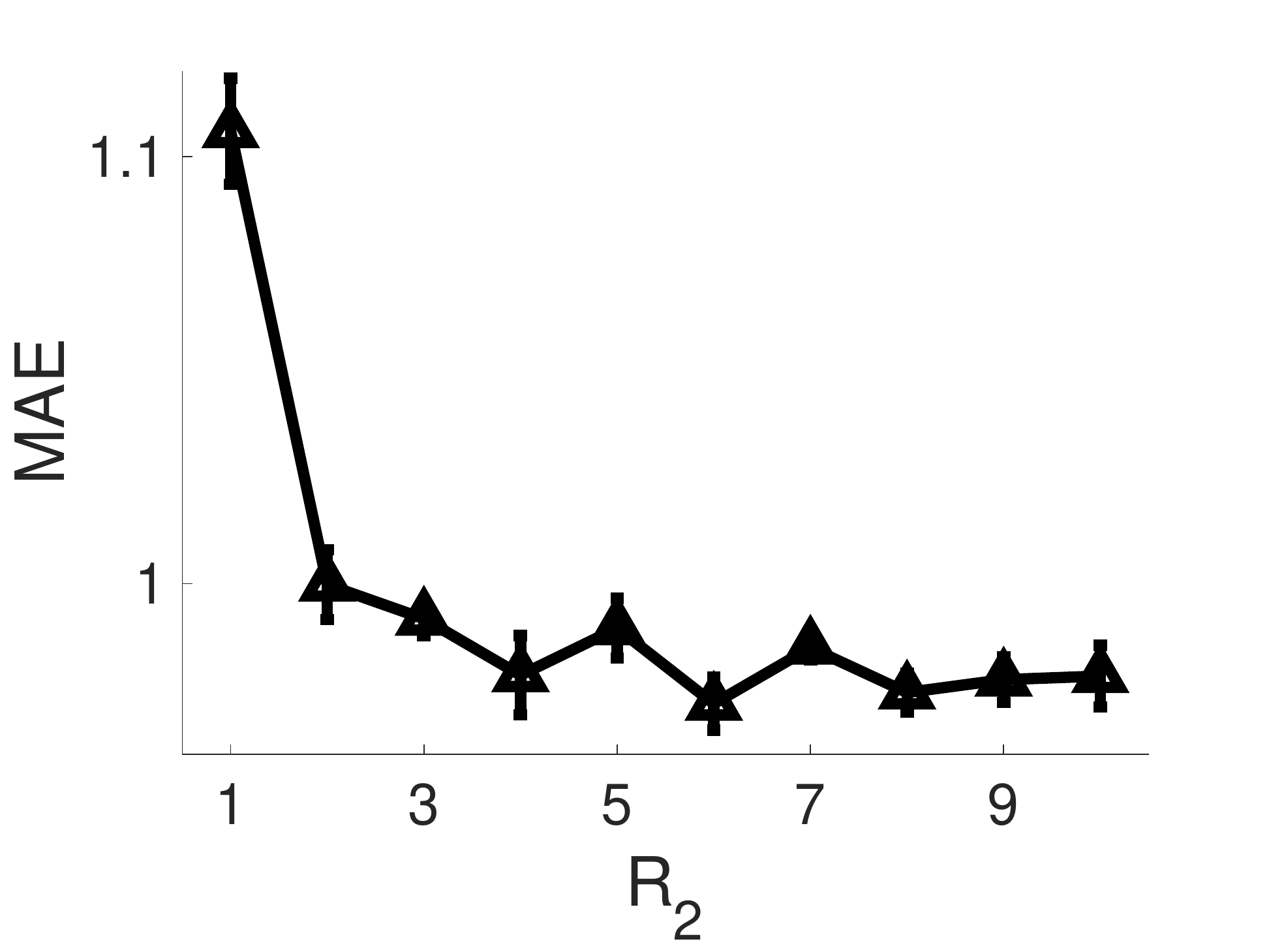}
			\caption{\small \textit{Alog}}
		\end{subfigure}
		&
		\begin{subfigure}[t]{0.25\textwidth}
			\centering
			\includegraphics[width=\textwidth]{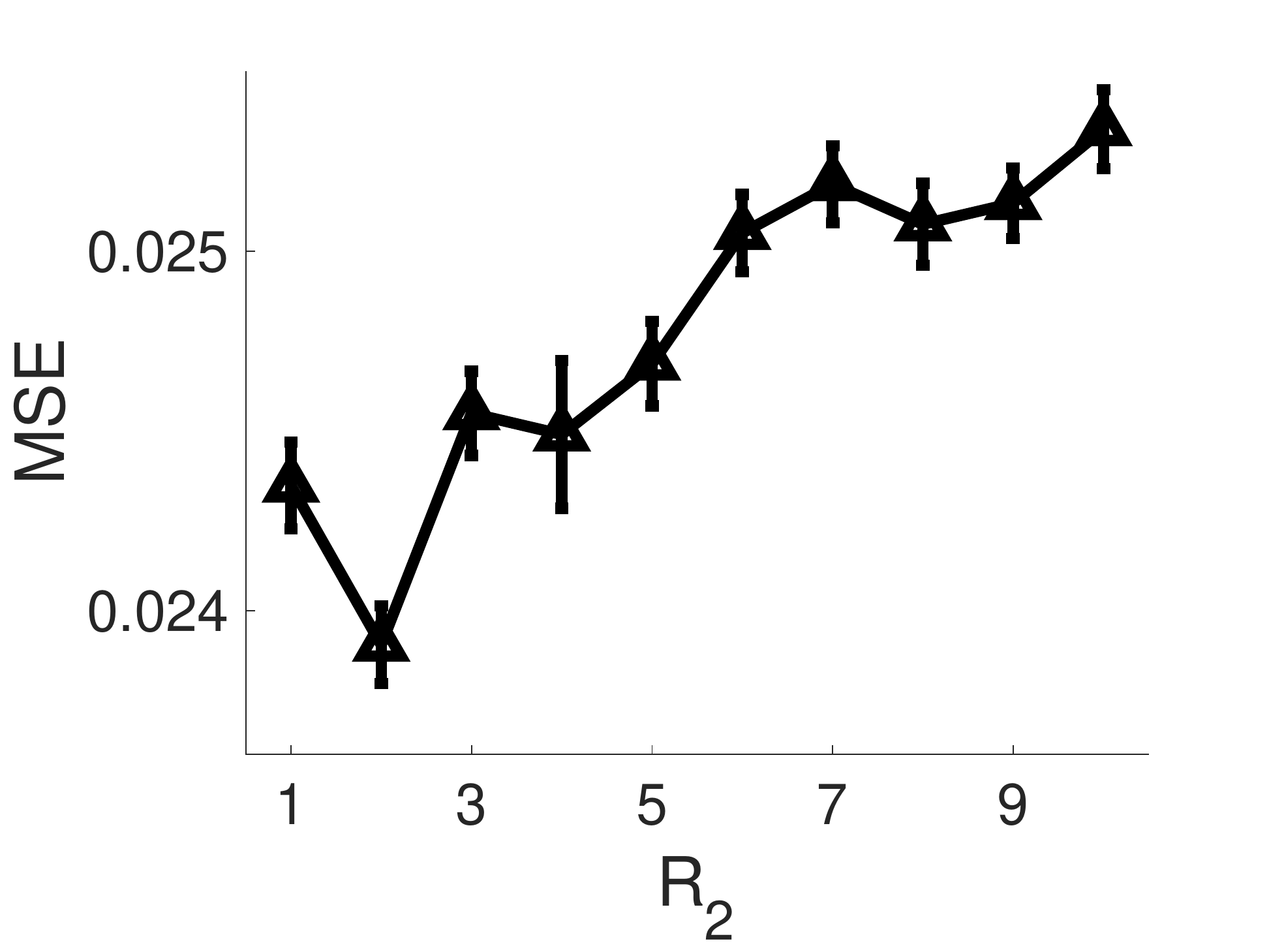}
			\caption{\small \textit{SG}}
		\end{subfigure}
		&
		\begin{subfigure}[t]{0.25\textwidth}
			\centering
			\includegraphics[width=\textwidth]{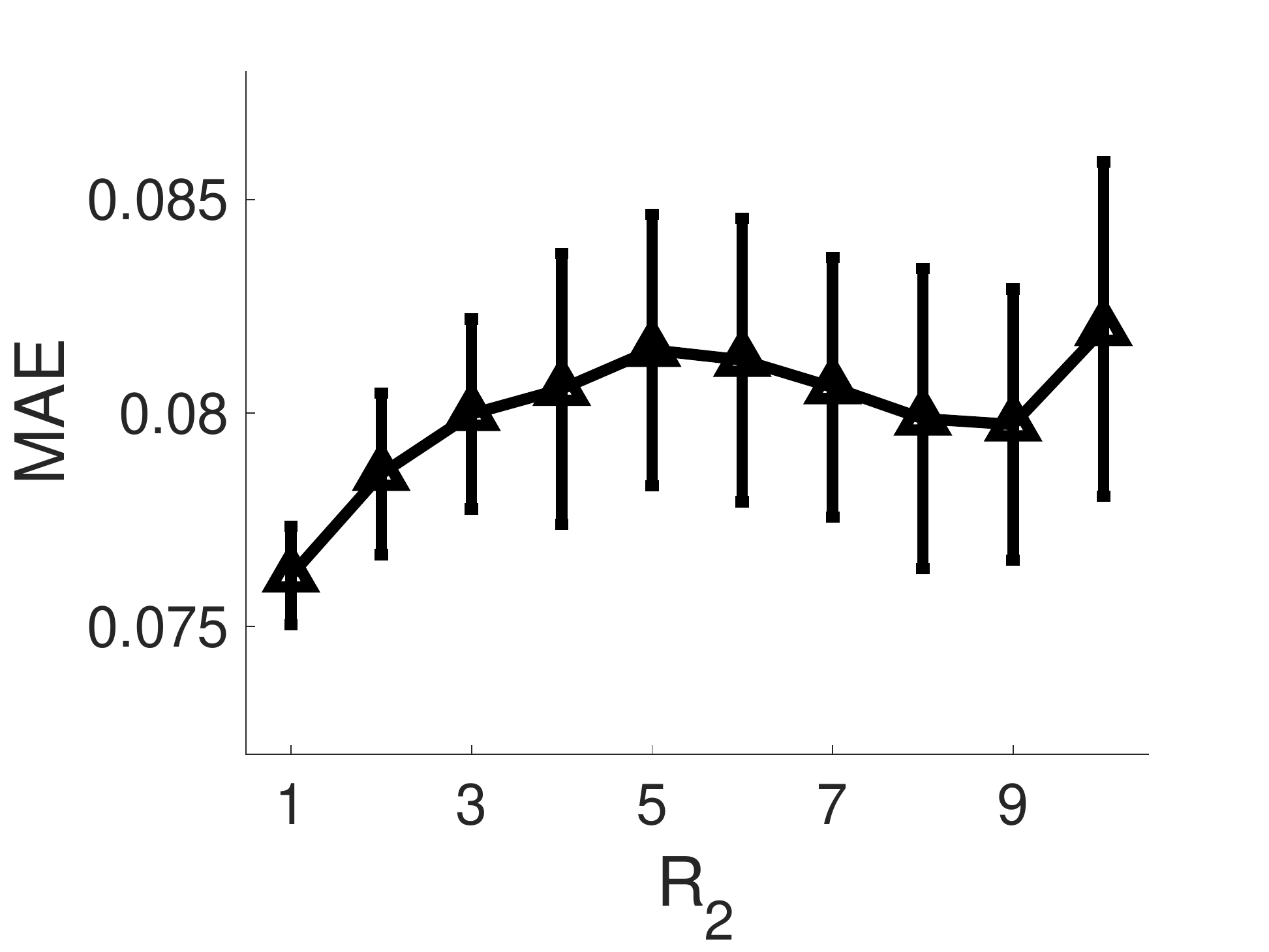}
			\caption{\small \textit{SG}}
		\end{subfigure}
	\end{tabular}
	\caption{\small The performance of our approach with different combinations of $R_1$ and $R_2$ (the number of location and sociability factors respectively). The total number of factors $R$ is fixed to $11$ ($R = R_1 + R_2$). }  
	\label{fig:comb}
\end{figure*}
 \begin{figure*}[ht]
	\centering
	\setlength{\tabcolsep}{0pt}
	\captionsetup[subfigure]{aboveskip=0pt,belowskip=0pt}
	\begin{tabular}[c]{ccc}
		\begin{subfigure}[t]{0.33\textwidth}
			\centering
			\includegraphics[width=\textwidth]{./figures/alog-mode-1-eps-converted-to.pdf}
			\caption{\textit{Alog} - Ours}
		\end{subfigure}
		&
		\begin{subfigure}[t]{0.33\textwidth}
			\centering
			\includegraphics[width=\textwidth]{./figures/movie-mode-1-eps-converted-to.pdf}
			\caption{\textit{MovieLens} - Ours}
		\end{subfigure}	 
		&
		\begin{subfigure}[t]{0.33\textwidth}
			\centering
			\includegraphics[width=\textwidth]{./figures/sg-mode-2-eps-converted-to.pdf}
			\caption{\textit{SG}- Ours}
		\end{subfigure}
	\\
	\begin{subfigure}[t]{0.33\textwidth}
		\centering
		\includegraphics[width=\textwidth]{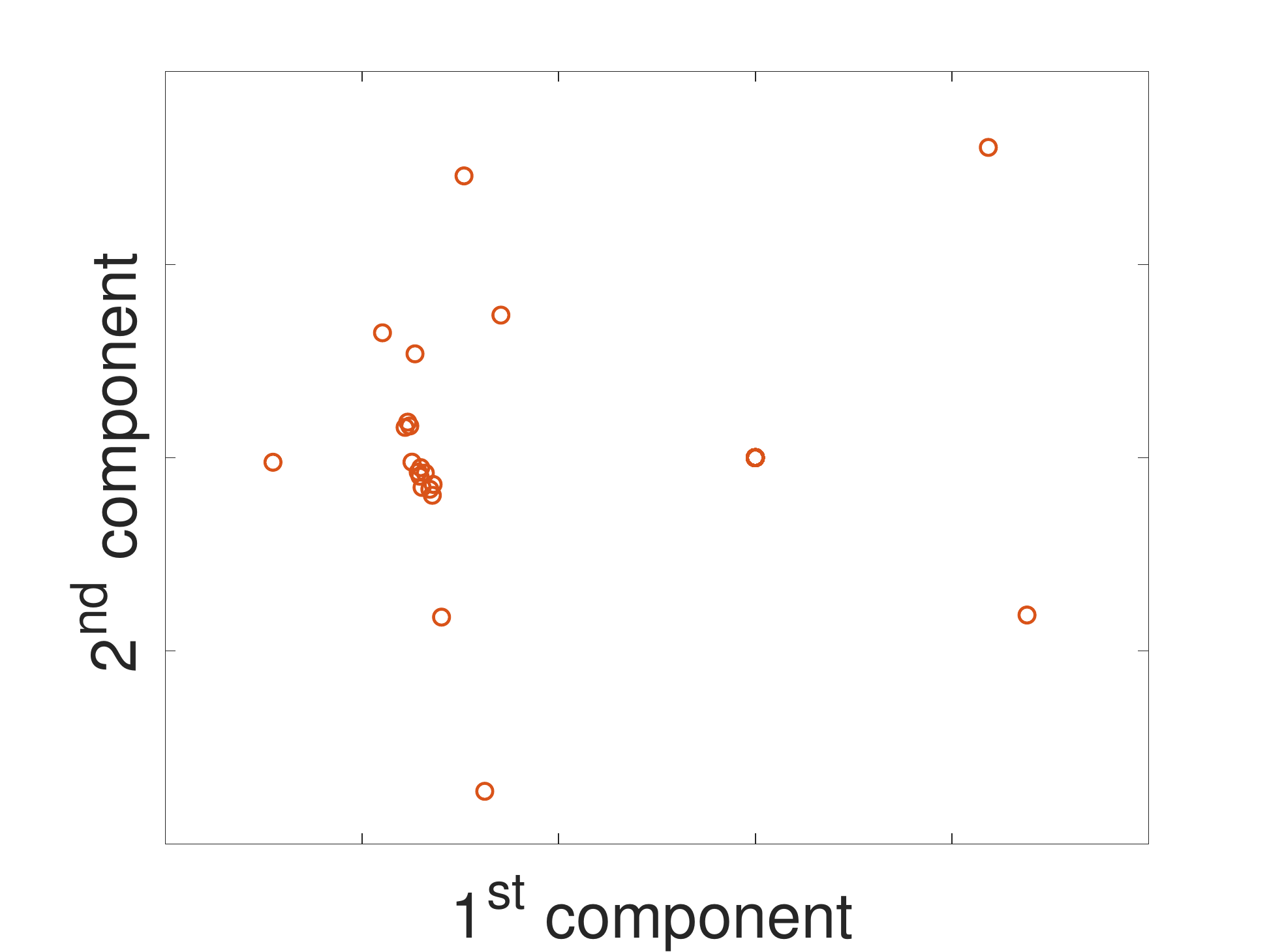}
		\caption{\textit{Alog} - CP-Bayes}
	\end{subfigure}
	&
	\begin{subfigure}[t]{0.33\textwidth}
		\centering
		\includegraphics[width=\textwidth]{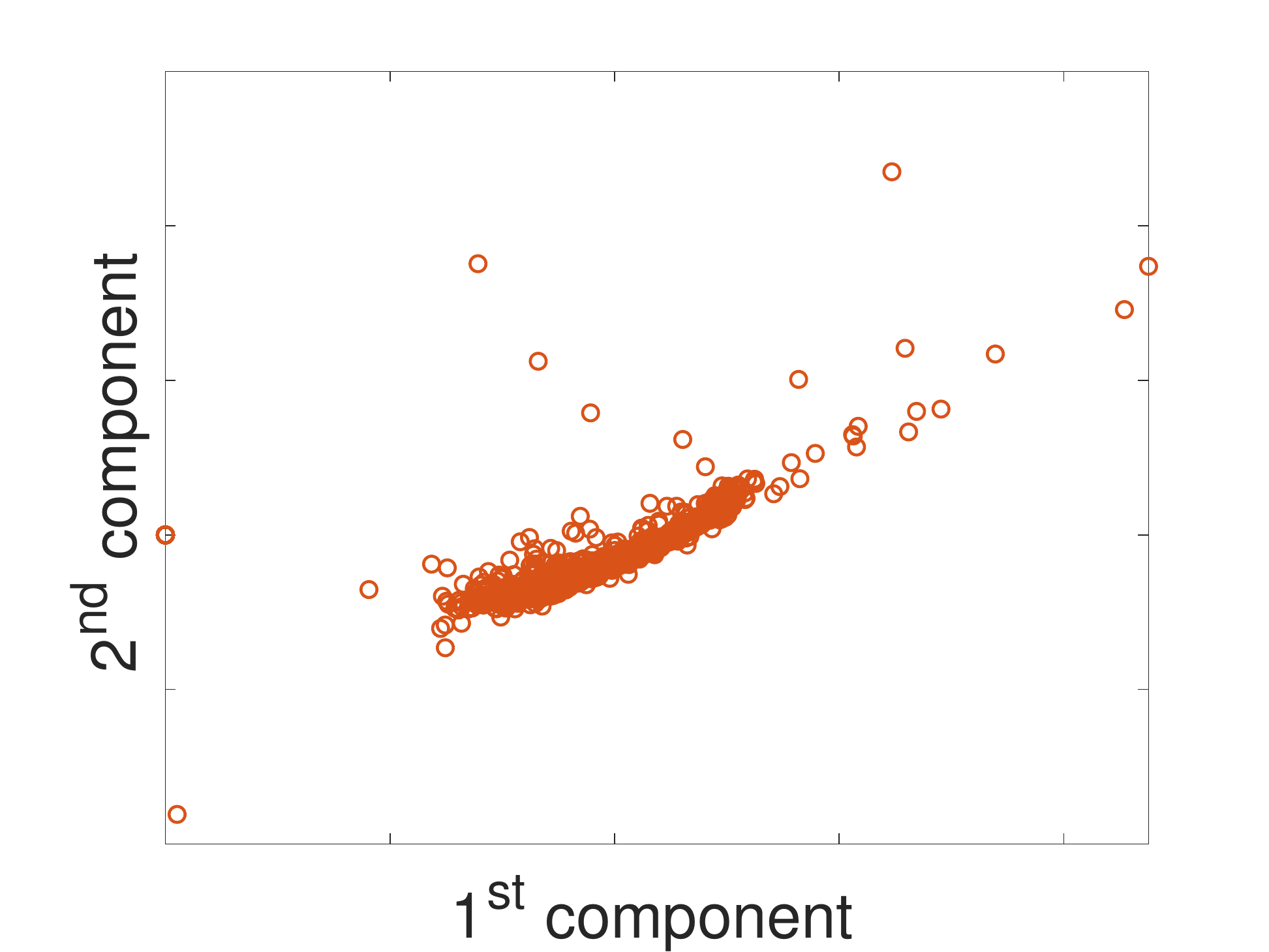}
		\caption{\textit{MovieLens} - CP-Bayes}
	\end{subfigure}	 
	&
	\begin{subfigure}[t]{0.33\textwidth}
		\centering
		\includegraphics[width=\textwidth]{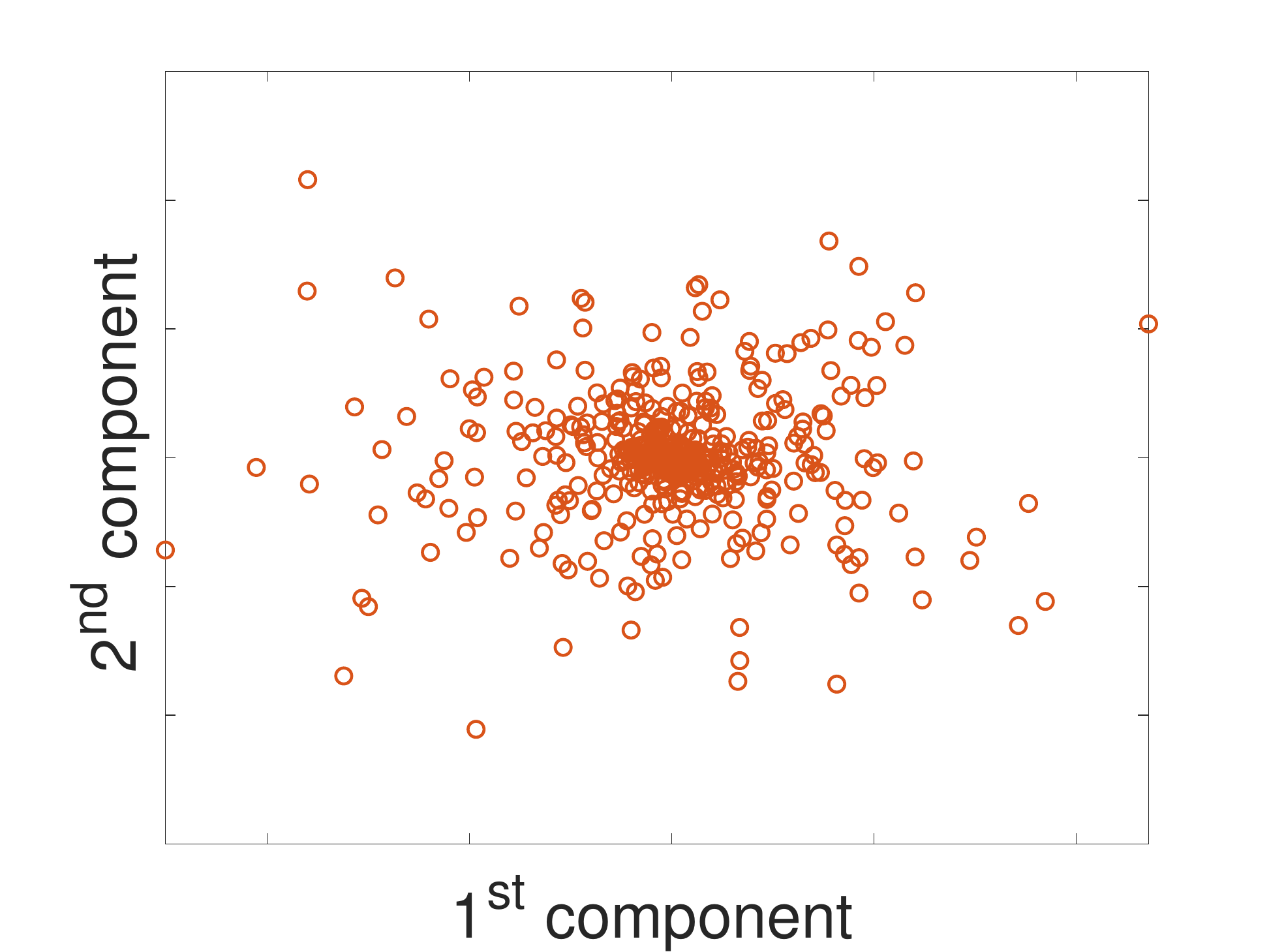}
		\caption{\textit{SG}- CP-Bayes}
	\end{subfigure}
	\\
		\begin{subfigure}[t]{0.33\textwidth}
		\centering
		\includegraphics[width=\textwidth]{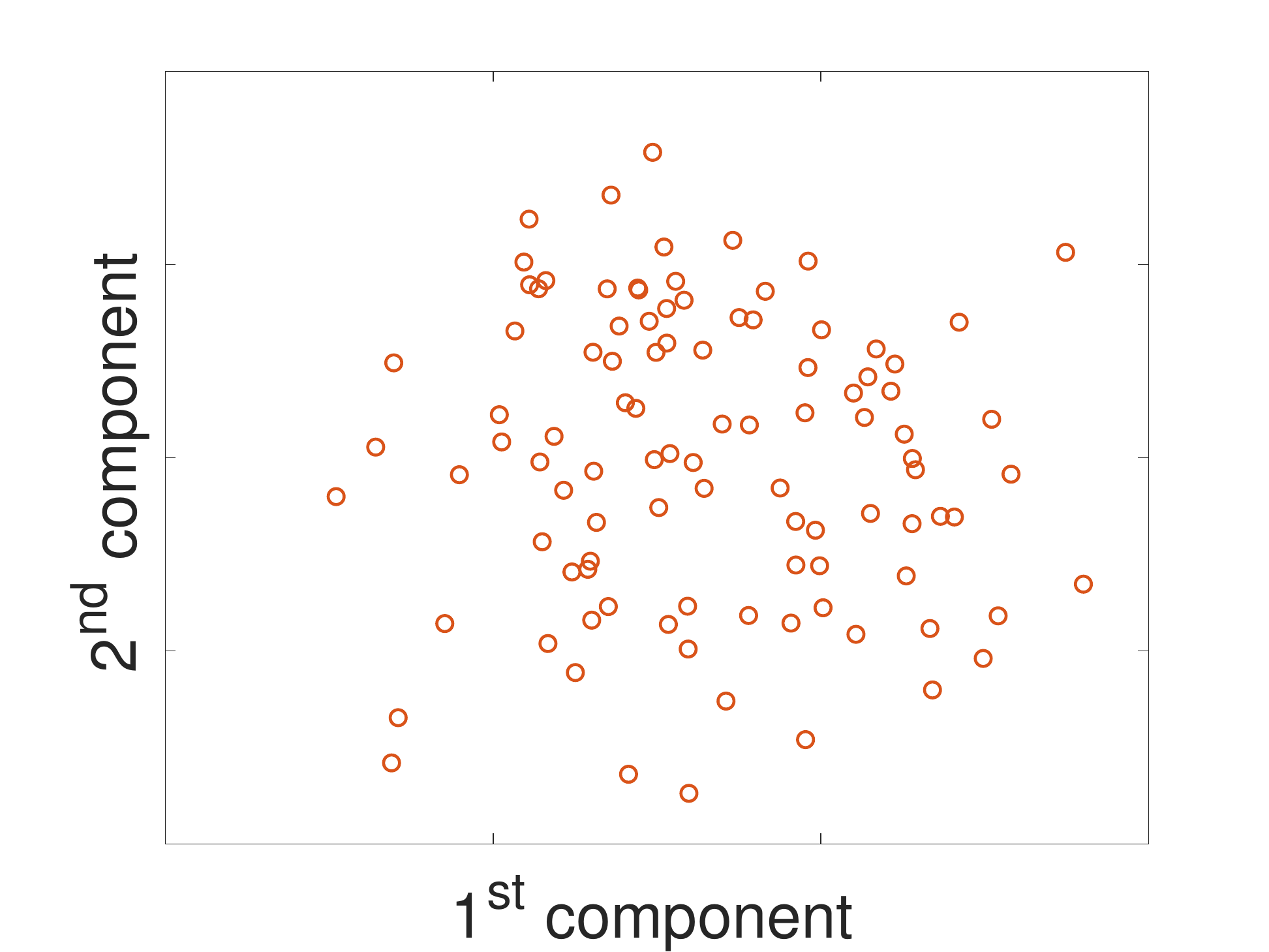}
		\caption{\textit{Alog} - GPTF}
	\end{subfigure}
	&
	\begin{subfigure}[t]{0.33\textwidth}
		\centering
		\includegraphics[width=\textwidth]{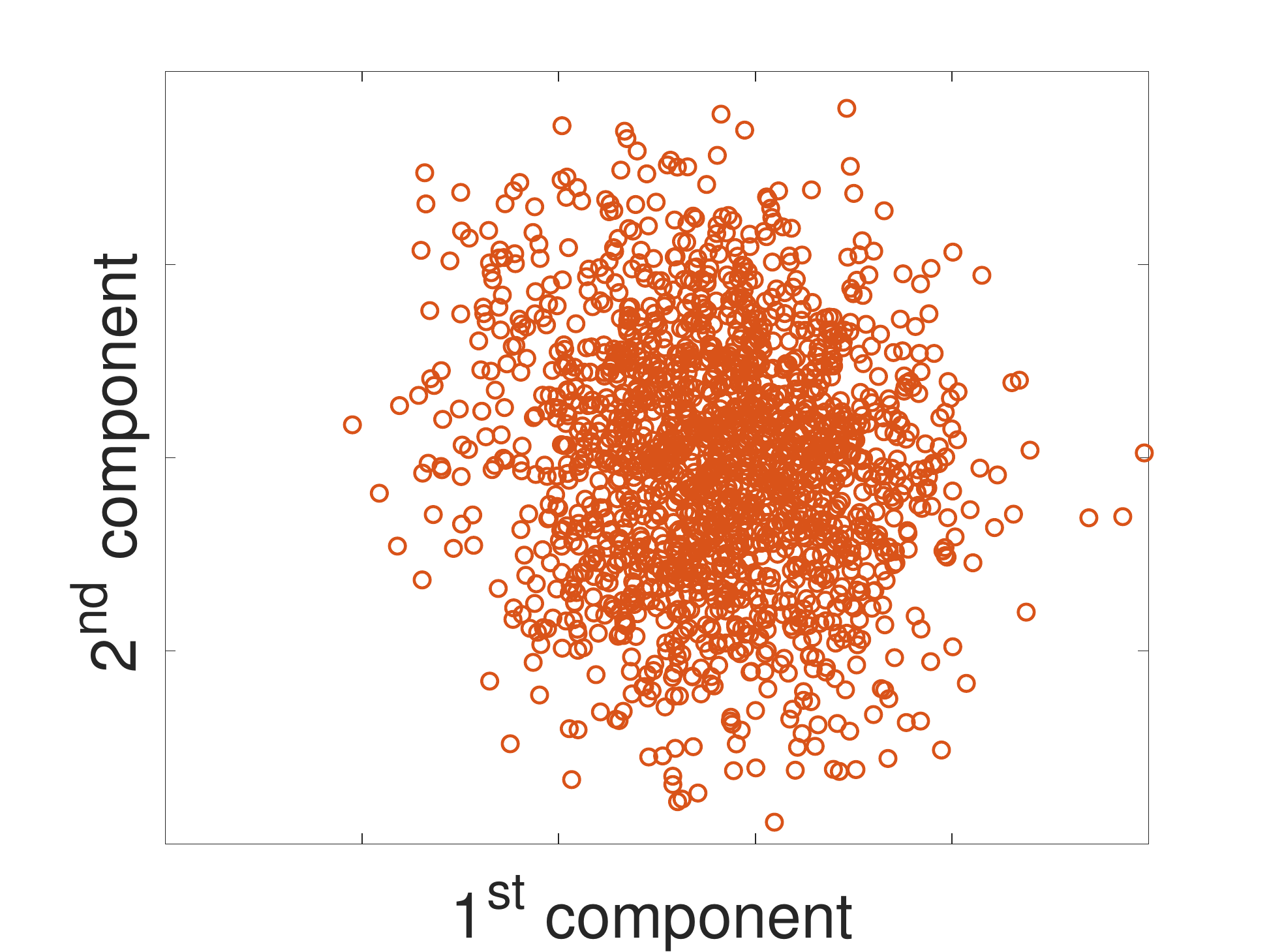}
		\caption{\textit{MovieLens} -GPTF}
	\end{subfigure}	 
	&
	\begin{subfigure}[t]{0.33\textwidth}
		\centering
		\includegraphics[width=\textwidth]{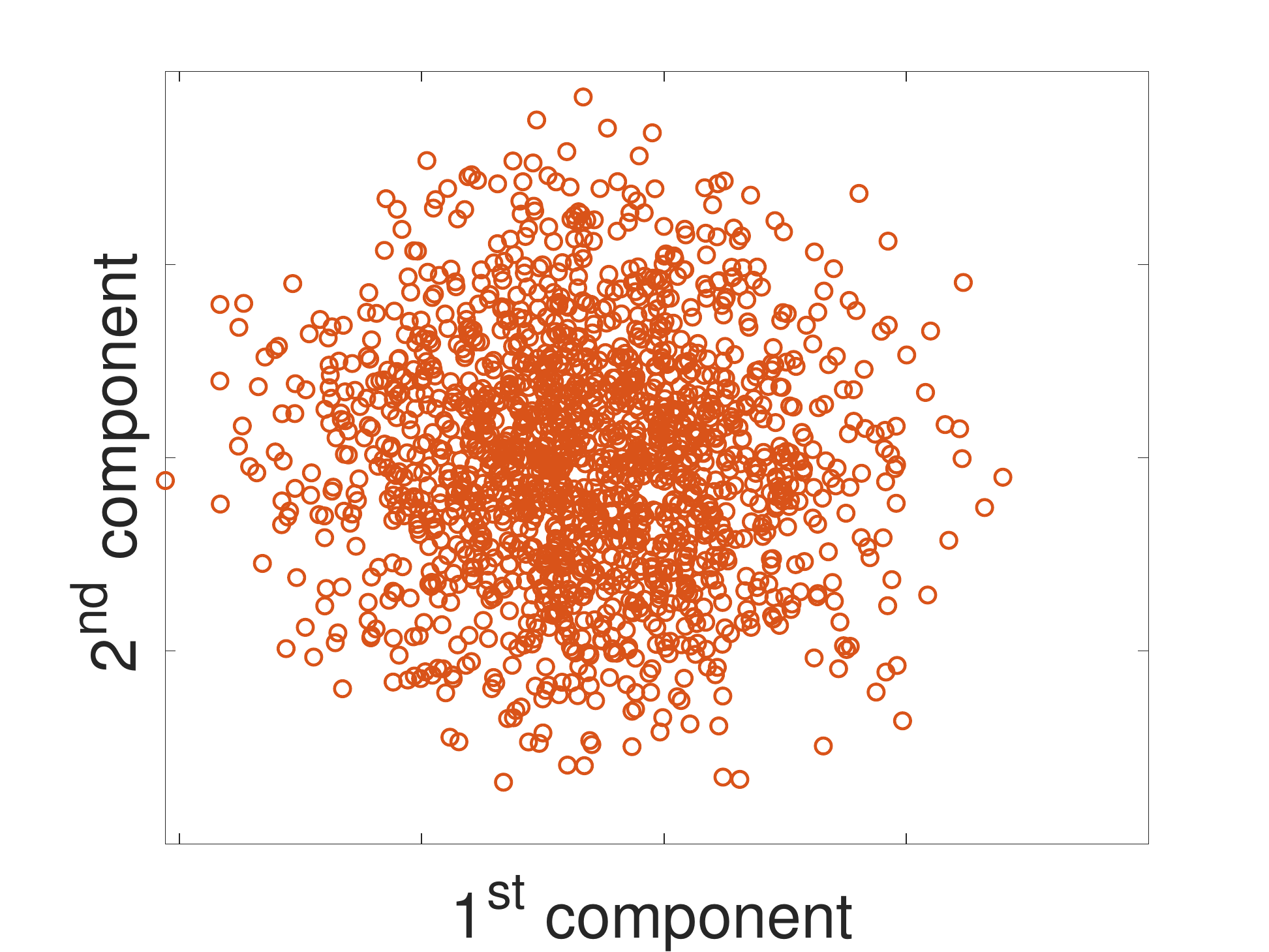}
		\caption{\textit{SG}- GPTF}
	\end{subfigure}
	\end{tabular}
	\caption{\small Structures of the estimated latent factors for the second mode of \textit{Alog} and \textit{MovieLens} dataset and third mode of \textit{SG} dataset, by our method, CP-Bayes and GPTF.} 
	\label{fig:pred-acc}
	\vspace{-0.2in}
\end{figure*}

\section{Combination of $R_1$ and $R_2$}
We investigated the performance of our method with different combinations of $R_1$ and $R_2$ (the number of location and sociability factors) on \textit{Alog} and \textit{SG} datasets. To this end, we fixed the total number of factors to $11$ and varied $R_2$, \ie the number of sociability factors. We used the same data splits as in Sec. 6.2 of the main paper. The average MSE and MAE for different combinations of $R_1$ and $R_2$ are reported in Fig. \ref{fig:comb}. We can see that the best setting on \textit{Alog} is $R_2 =  6, R_1 = 5$, and on \textit{SC} is $R_2 = 2, R_1 = 9$ (MSE) and $R_2 = 1, R_1 =10$ (MAE). While the results of all the settings are better than all the competing dense tensor models, different combinations of $R_1$ and $R_2$ result in quite different prediction accuracies. The best trade-off is application specific. The results also demonstrate the advantage of our method over NEST models~\citep{tillinghast2021nonparametric}, which restrict the types of factors to be estimated and do not have the flexibility to choose different $R_1$ and $R_2$.

\section{Running Time}
The speed of our method is faster than NEST, comparable to GPTF, and slower than the other competing methods. For example, for $R=5$, on \textit{MovieLens} and \textit{SG} datasets, the average running time (in seconds) of each method is \{Ours: 825.1, GPTF: 706.5, NEST-1: 1218.9, NEST-2: 1193.4, CP-Bayes: 323.2, P-Tucker: 7.9, CP-WOPT: 6.1, CP-ALS: 0.8\} and  \{Ours: 1213.5, GPTF: 1096.1, NEST-1: 2112.2, NEST-2: 1910.5, CP-Bayes: 566.4, P-Tucker: 10.8, CP-WOPT: 10.6, CP-ALS: 0.7\}. This is reasonable, because the GP based methods, although using sparse approximations to avoid computing the full covariance (kernel) matrix, still need to estimate the frequencies and perform  extra nonlinear feature transformations (\ie random Fourier features). Hence, they are much more complex. In addition, our method introduces nonparametric priors over sparse tensor entries, and the estimation involves discrete, hierarchical probability measures. Therefore, it requires more computation.

\section{Proof of Lemma 3.1 and 3.2}

\newcommand{\floor}[1]{\lfloor #1 \rfloor}
\subsection{Proof of Lemma 3.1}\label{sect:lem31}

To ease the notation, we consider the case where $R_1=R_2=1$. It is straightforward to extend the proof to arbitrary $R_1$ and $R_2$.  Given a particular $\alpha < \infty$, denote the total number of points in $T$ (see (4) in the main paper) by $D^\alpha$. We observe that by the properties of the Poisson random measure, $D^\alpha \sim \text{Poisson}(W_1^\alpha([0,\alpha])\times \dots\times W_K^\alpha([0,\alpha]))$.  Since these points might overlap, the  number of the distinct points (\ie entries)  $N^\alpha \le D^\alpha$. According to the definition of Gamma process,  $W_k^\alpha([0, \alpha]) \vert L_k^\alpha$ is a Gamma distributed random variable with mean given by $L_k^\alpha([0, \alpha])$, and $L_k^\alpha([0, \alpha])$ is another Gamma random variable with mean $\alpha$. Therefore, $L_k^\alpha([0, \alpha]) < \infty$  almost surely, $W_k^\alpha([0, \alpha]) \vert L_k^\alpha < \infty $ almost surely, and then $W_k^\alpha([0, \alpha]) < \infty$ almost surely (because $p(W_k^\alpha([0, \alpha])< \infty) = \int p\left(W_k^\alpha([0, \alpha])< \infty|L_k^\alpha\right)  \d p(L_k^\alpha) = 1$). Hence, their product  $b^\alpha = W_1^\alpha([0,\alpha])\times \dots\times W_K^\alpha([0,\alpha])$ is finite almost surely. Since $D^\alpha \sim \text{Poisson}(b^\alpha)$,  we have $D^\alpha < \infty$ almost surely and $N^\alpha \le D^\alpha$ implies  $N^\alpha < \infty$ almost surely.

Now we consider $W_k^\alpha((j-1,j])\vert L_k^\alpha$ where $j \in \mathbb{N}^+$. This is a  Gamma distributed random variable with mean given by $L_k^\alpha((j-1,j])$. However $L_k^\alpha((j-1,j])$ is a mean-one Gamma distributed random variable. Thus if we integrate out $L^\alpha_k$ for  $1\le k\le K $, $\{W^\alpha_k((j-1,j])\}_{k,j}$ are independent identically distributed random variables because $L^\alpha_k$ as a Gamma process itself is independent on disjoint sets. Note that given $j$, $\{W^\alpha_k((j-1,j])\}$ across $k$ have already been independent. Furthermore, we have $\EE[W_k^\alpha((j-1,j])] = 1$, and  $\EE[W_1^\alpha((j-1,j]) \times \ldots \times W_K^\alpha((j-1, j])] = 1$. 

We consider $S_{\floor{\alpha}} = \sum_{j=1}^{\floor{\alpha}} \mathds{1}(X_j>0)$ where $\mathds{1}(\cdot)$ is the indicator function, and $$X_j \sim \text{Poisson}\left(W_1^\alpha((j-1, j]) \times \ldots \times  W_K^\alpha((j-1, j])\right).$$ Obviously, $S_{\floor{\alpha}} \le N^\alpha$. For convenience,  let us define $\zeta_j = W_1^\alpha((j-1, j]) \times \ldots \times  W_K^\alpha((j-1, j])$. We have $\EE[X_j|\zeta_j] = \zeta_j$ and $\EE[X_j] = \EE[\zeta_j] = 1$ .   Since all $\{X_j\}_j$ are i.i.d,  the indicators $\{\mathds{1}(X_j>0)\}_j$ are i.i.d as well. According to the strong law of large numbers, $$\lim_{\floor{\alpha} \rightarrow \infty} \frac{S_{\floor{\alpha}}}{\floor{\alpha}} = \EE[\mathds{1}(X_j>0)] = 1-e^{-1}>0 \,\, a.s.$$ Therefore, when $\alpha \rightarrow \infty$, $\floor{\alpha} \rightarrow \infty$ and 
$S_{\floor{\alpha}} \rightarrow \infty$ a.s. Since $N^\alpha \ge S_{\floor{\alpha}}$, we have $\lim_{\alpha \rightarrow \infty}N^{{\alpha}} = \infty$ a.s.

\subsection{Proof of Lemma 3.2}
Our tensor-variate stochastic process when $R_2 = 1$ is summarized in the following.
\begin{align*}
    L^\alpha_k & \sim \Gamma \text{P}(\lambda_\alpha),\\
    W^\alpha_k \mid L^\alpha_k & \sim \Gamma \text{P}(L^\alpha_k),\\  
    T \mid \{W^\alpha_k\}_{k=1}^\infty& \sim  
    \text{PRM}(W^\alpha_1 \times \dots \times W^\alpha_K).  
\end{align*}

We will follow the similar strategy in \citep{tillinghast2021nonparametric} to prove the lemma. That is, given $\alpha$, we define $M^\alpha_k$ the number of active nodes in mode $k$ and  $N^\alpha$ the number of sampled entries. Then we have $M^\alpha_k = \#\{\theta^{k}_i\in[0,\alpha] \vert T(A^\alpha_{k,\theta_i^k})>0 \}$, where $A^\alpha_{k,\theta_i^k} = [0,\alpha]\times\dots \times  \{\theta^{k}_i\}\times\dots\times [0,\alpha]$. The goal of the first step is to show $\lim_{\alpha\rightarrow \infty} \frac{\alpha}{M_{k}^\alpha} = 0 \;\; a.s.$ for all $k\in \{1, \dots , K \}$, and the second step is $\limsup_{\alpha \rightarrow \infty}N^\alpha/\alpha^K<\infty\;\; a.s$. Based on these, it is trivial to show  $$\lim_{\alpha\rightarrow\infty } \dfrac{N^\alpha}{\prod_{k=1}^K M_k^\alpha} = 0\;\; a.s.$$ 

\noindent \textbf{Step 1.} Following the same steps as in \citep{tillinghast2021nonparametric}, we can show that 
\begin{align*}
	&\EE[M_k^\alpha\vert\{W_j^\infty\}_{j\neq k},L_k^\alpha] \notag \\
	& =\EE\left[\sum_{\theta_i^k\in[0,\alpha]} 1-\exp\left(-W_k^\infty(\{\theta^{k}_i\}\times \prod_{j\neq k}W^\infty_j([0,\alpha])\right)  \bigg\vert \{W^\infty_j\}_{j\neq k},L_k^\alpha  \right]. 
\end{align*}
Note that different from \citep{tillinghast2021nonparametric}, the expectation here is conditioned on $L_k^\alpha$, an additional level of \gaps. We then apply Campbell's Theorem~\citep{kingman1992poisson} and obtain
\begin{align*}
	&\EE[M_k^\alpha\vert\{W_j^\infty([0,\alpha])\}_{j\neq k},L_k^\alpha]  \\
	& = L_k^\alpha([0,\alpha]) \int_0^\infty\left( 1-\exp\left(-w\times \prod_{i\neq k}W^\infty_i([0,\alpha])\right)\right)w^{-1}e^{-w}dw.
\end{align*}
Since $L_k^\alpha([0,\alpha])$ is a Gamma random variable with mean $\alpha$,  we further take the expectation with respect to the measure induced by $L_k^\alpha$. Then, we obtain 
$$ \EE[M_k^\alpha\vert\{W_j^\infty([0,\alpha])\}_{j\neq k}] = \alpha\cdot \int_0^\infty\left( 1-\exp\left(-w\times \prod_{i\neq k}W^\infty_i([0,\alpha])\right)\right)w^{-1}e^{-w}dw.$$
Now, we arrive at the same intermediate result as in \citep{tillinghast2021nonparametric}. It follows that $$ \lim_{\alpha\rightarrow \infty} \frac{\alpha}{M_{k}^\alpha} = 0 \;\; a.s.$$
\textit{The result points out that in each mode, the number of active nodes  $M_k^\alpha$ grows faster than $\alpha$}.

\noindent \textbf{Step 2.}  We consider the distribution of $W_k([j,j+1])\mid L_k^\alpha$. This will be Gamma distributed with mean $L_k^\alpha([j,j+1])$. But $L_k^\alpha(j,j+1)$ is Gamma distributed with mean 1 for all $j<\alpha-1$. Thus by marginalizing out $L_k^\alpha(j,j+1)$ we see that $W_k([j,j+1])$ is identically distributed for all $j<\alpha-1$.
By the independence of the completely random measure~\citep{kingman1967completely} on disjoint sets, it follows immediately by the strong law of large numbers $$\lim_{j\rightarrow \infty}\dfrac{ W_k^\infty([0,j])}{j} =\dfrac{ \sum_{i=1}^j W_k^\infty((i-1,i])}{j} =E[W_k^\infty([0,1])] = 1\, a.s.$$ as $W_k^\infty((i-1,i])$ are $\text{i.i.d}$ random variables. This implies \begin{equation}\lim_{j\rightarrow \infty}\dfrac{\prod_{k=1}^K W_k^\infty([0,j])}{j^K} = 1\, a.s. \label{eq:limit-2}\end{equation}

Let us denote $D^\alpha$ by the actual number of points sampled in $T([0, \alpha]^K)$. Note $N^\alpha \le D^\alpha$. Applying Lemma 17 in \citep{caron2014sparse} implies  
$$Pr\left(\lim_{j\rightarrow\infty}\dfrac{D^j}{\prod_{k=1}^K W_k^\infty([0,j])} = 1\bigg\vert \{W^\infty_i\}_{i= 1}^K\right)=1.$$
Taking the expectation of both sides of the above expression and combining with equation \eqref{eq:limit-2} implies $$\lim_{j\rightarrow\infty}\dfrac{D^j}{j^K} = 1\, a.s.$$ Now, we can use the same bounding technique as in \citep{tillinghast2021nonparametric} to show that $$\lim_{\alpha \rightarrow \infty}\dfrac{N^\alpha}{\alpha^K} \le \lim_{\alpha \rightarrow \infty}\dfrac{D^\alpha}{\alpha^K} = 1 < \infty.$$  \textit{The result in the second step points out the number of sampled entries $N^\alpha$ grows slower than or as fast as $\alpha^K$}.

Finally, we extend the case when $R_2>1$ by simply applying Poisson superposition theorem.

\end{document}